# Generating Reflectance Curves from sRGB Triplets

Scott Allen Burns
University of Illinois at Urbana-Champaign (scottb@illinois.edu)

## Overview

I present several algorithms for generating a reflectance curve from a specified sRGB triplet, written for a general audience. Although there are an infinite number of reflectance curves that can give rise to the specific color sensation associated with an sRGB triplet, the algorithms presented here are designed to generate reflectance curves that are similar to those found with naturally occurring colored objects. My hypothesis is that the reflectance curve with the least sum of slope squared (or in the continuous case, the integral of the squared first derivative) will do this. After presenting the algorithms, I examine the quality of the computed reflectance curves compared to thousands of documented reflectance curves measured from paints and pigments available commercially or in nature. Being able to generate reflectance curves from three-dimensional color information is useful in computer graphics, particularly when modeling color transformations that are wavelength specific.

## Introduction

There are many different 3D color space models, such as XYZ, RGB, HSV, L*a*b*, etc., and one thing they all have in common is that they require only three quantities to describe a unique color sensation. This reflects the "trichromatic" nature of human color perception. The space of *color stimuli*, however, is not three dimensional. To specify a unique color stimulus that enters the eye, the power level at every wavelength over the visible range (e.g., 380 nm to 730 nm) must be specified. Numerically, this is accomplished by discretizing the spectrum into narrow wavelength bands (e.g., 10 nm bands), and specifying the total power in each band. In the case of 10 nm bands between 380 and 730 nm, the space of color stimuli is 36 dimensional. As a result, there are many different color stimuli that give rise to the same color sensation (infinitely many, in fact).

For most color-related applications, the three-dimensional representation of color is efficient and appropriate. But it is sometimes necessary to have the full wavelength-based description of a color, for example, when modeling color transformations that are wavelength specific, such as dispersion or scattering of light, or the subtractive mixture of colors, for example, when mixing paints or illuminating colored objects with various illuminants. In fact, this document was developed in support of another document concerning how to compute the RGB color produced by subtractive mixture of two RGB colors.[1]

I present several algorithms for converting a three-dimensional color specifier (sRGB) into a wavelength-based color specifier, expressed in the form of a reflectance curve. When quantifying object colors, the reflectance curve describes the fraction of light that is reflected from the object by wavelength, across the visible spectrum. This provides a convenient, dimensionless color specification, a curve that varies between zero and one (although fluorescent objects can have reflectance values >1). The motivating idea behind these algorithms is that the one reflectance curve that has the least sum of slope squared (integral of the first derivative squared, in the continuous case) seems to match reasonably well the reflectance curves measured from real paints and pigments available commercially and in nature. After presenting the algorithms, I compare the computed reflectance curves to thousands of documented reflectance measurements of paints and pigments to demonstrate the quality of the match.

**sRGB Triplet from a Reflectance Curve**

The reverse process of computing an sRGB triplet from a reflectance curve is straightforward. It requires two main components: (1) a mathematical model of the "standard observer," which is an empirical mathematical relationship between color stimuli and color sensation (tristimulus values), and (2) a definition of the sRGB color space that specifies the reference illuminant and the mathematical transformation between tristimulus values and sRGB values.

The linear transformation relating a color stimulus, $N$, to its corresponding tristimulus values, $Q$, is

$$Q_{3\text{x}1} = A'_{3\text{x}n}\, N_{n\text{x}1}.$$

The column vector $Q$ has three elements, $X$, $Y$, and $Z$. The matrix $A'$ has three rows (called the three "color matching functions") and $n$ columns, where $n$ is the number of discretized wavelength bands. In this study, all computations are performed with 36 wavelength bands of width 10 nm, running over the range 380 nm to 730 nm. The stimulus vector also has $n$ components, representing the total power of all wavelengths within each band. The specific color matching functions I use in this work are the CIE 1931 color matching functions.[2] (Note that I'm using the symbol $A'$ here; the standard $A$ matrix is $n$ x 3, and so $A'$ indicates that it has been transposed.)

The stimulus vector can be constructed as the product of an $n\text{x}n$ diagonal illuminant matrix, $\text{diag}(W)$, and an $n\text{x}1$ reflectance vector, $\rho$. The computation of $Q$ is usually normalized so the $Y$ tristimulus value is 1 when $\rho$ is a perfect reflector (contains all 1s). The normalizing factor, $w$, is thus the inner product of the second row of $A'$ and the illuminant vector, $W$, yielding the alternate form of the tristimulus value equation

$$Q = A'\, \text{diag}(W)\, \rho / w.$$

The transformation from tristimulus values to sRGB is a two-step process. First, a 3x3 linear transformation, $M^{-1}$, is applied to convert $Q$ to $rgb$, which is a triplet of "linear RGB" values:

$$rgb = M^{-1}\, A'\, \text{diag}(W)\, \rho / w.$$

The second step is to apply a "gamma correction" to the linear $rgb$ values, also known as "companding" or applying the "color component transfer function." This is how it is done: for each $r$, $g$, and $b$ component of $rgb$, let's generically call it $v$, if $v \leq 0.0031308$, use $12.92v$, otherwise use $1.055v^{1/2.4} - 0.055$. This gives sRGB values in the range of 0 to 1. To convert them to the alternate integer range of 0 to 255 (used in most 24-bit color devices), we multiply each by 255 and round to the nearest integer.

The inverse operation of converting sRGB to $rgb$, expressed in (Matlab) code, is:

```
sRGB=sRGB/255; % convert 0-255 range to 0-1 range
for i=1:3
  if sRGB(i)<0.04045
    rgb(i)=sRGB(i)/12.92;
  else
    rgb(i)=((sRGB(i)+0.055)/1.055)^2.4;
  end
end
```

The expression relating $rgb$ and $\rho$ above can be simplified by combining the three matrices and the normalizing factor into a single matrix,

$$T = M^{-1}\ A'\ \mathrm{diag}(W)/w$$

so that

$$rgb = T\ \rho.$$

The formal definition[3] of the sRGB color space uses an illuminant similar to daylight, called D65, as its "reference" illuminant. Here are the specific values for the $M^{-1}$ matrix, the $A'$ matrix, the D65 $W$ vector, and the $T$ matrix. The normalizing factor, $w$, has a value of 10.5677. Most of the RGB-related theory I present here comes from Bruce Lindbloom's highly informative website.[4]

Now that we have a simple expression for computing sRGB from a reflectance curve, we can use that as the basis of doing the opposite, computing a reflectance curve from an sRGB triplet. In the sections that follow, I will present five different algorithms for doing this. Each has its strengths and weaknesses. Once they are presented, I will then compare them to each other and to reflectance curves found in nature.

**Linear Least Squares (LLS) Method**

Since there are so many more columns of $T$ than rows, the linear system is under-determined and gives rise to an $n - 3$ dimensional subspace (33-dimensional in our case) of reflectance curves for a single sRGB triplet. There are well-established techniques for solving under-determined linear systems. The most common method goes by various names: the linear least squares method, the pseudo-inverse, the least-squares inverse, the Moore-Penrose inverse, and even the "Fundamental Color Space" fundamental metamer.[5]

Suppose we pose this optimization problem:

$$\text{minimize } \rho^{\mathsf{T}}\rho$$
$$\text{s.t. } T\rho = rgb$$

This linearly constrained minimization can be solved easily by forming the Lagrangian function

$$L(\rho, \lambda) = \rho^{\mathsf{T}}\rho + \lambda^{\mathsf{T}}(T\rho - rgb).$$

The solution can be found by finding a stationary point of $L$, i.e., setting partial derivatives with respect to $\rho$ and $\lambda$ equal to zero:

$$\partial L/\partial \rho = 2\rho + T^{\mathsf{T}}\lambda = 0$$
$$\partial L/\partial \lambda = T\rho - rgb = 0.$$

Solving this system by eliminating $\lambda$ gives the LLS solution

$$\rho = T^{\mathsf{T}}(T\,T^{\mathsf{T}})^{-1}\ rgb.$$

Thus, a reflectance curve can be found from a sRGB triplet by simply converting it to linear $rgb$ and multiplying it by a 3x3 matrix, $T^{\mathsf{T}}(T\,T^{\mathsf{T}})^{-1}$. Unfortunately, the resulting solution is sometimes not very useful. Consider its application to this reflectance curve, which represents a bright red object color:

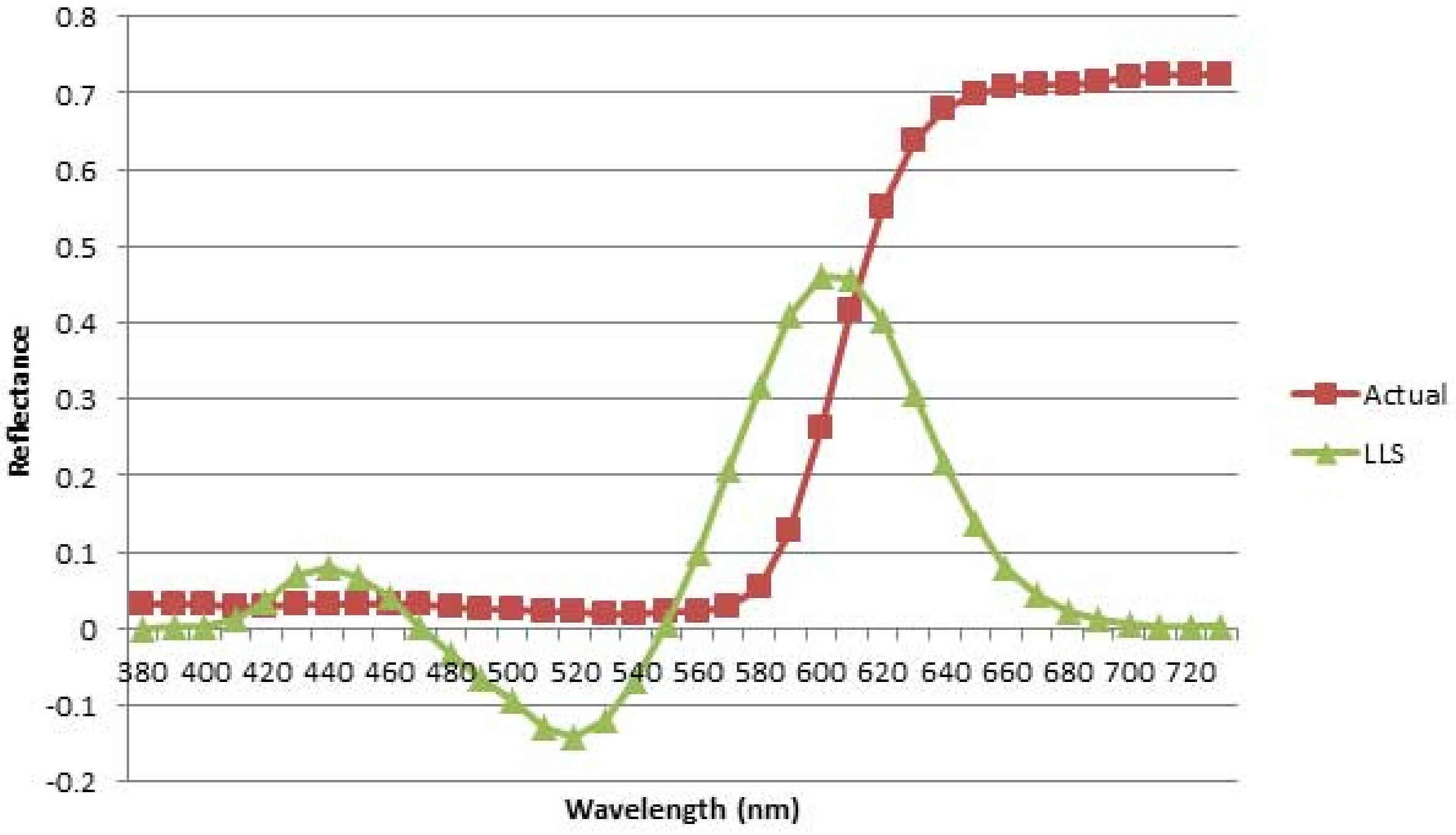

Reflectance curve for Munsell 5R 4/14 color sample and linear least-squares reconstruction.

The LLS solution contains negative reflectance values, which don't have physical meaning and limit its usefulness in realistic applications. Computationally, this is a very efficient method. The matrix $T^{\mathsf{T}}(T\ T^{\mathsf{T}})^{-1}$ can be computed in advance, as shown here, and each new sRGB value needs only be multiplied by it to get a reflectance curve.

**Least Slope Squared (LSS) Method**

Note that the standard LLS method minimizes $\rho^{\mathsf{T}}\rho$, that is, it finds the solution that is nearest to the origin, or the reflectance curve that oscillates most tightly about the wavelength axis. For purposes of computing reflectance curves, I can't think of a compelling reason why this should be a useful objective.

It dawned on me that it might be better to try a different objective function. The reflectance curves of most natural colored objects don't tend to oscillate up and down very much. I came up with the idea to minimize the square of the slope of the reflectance curve, summed over the entire curve. In the continuous case, this would be equivalent to

$$\min \int_{visible\ \lambda} (d\rho/d\lambda)^2\, d\lambda.$$

The square is used because it equally penalizes upward and downward movement of $\rho$. This objective will favor flatter reflectance curves and avoid curves that have a lot of up and down movement. (I later learned that this objective function has been previously investigated by C van Trigt in 1990 in a pair of publications.[5])

Other researchers have developed methods for reconstructing reflectance curves from tristimulus values. In order to reduce the oscillations, they typically introduce basis functions that oscillate very little, such as segments of low-frequency sinusoids, or they "frequency limit" or "band limit" the solution by constraining portions of the Fourier transform of the reflectance curve. To my mind, these approaches seem to ignore that fact that realistic reflectance curves can sometimes exhibit sudden steep changes in reflectance at certain frequencies, which would have relatively large high frequency Fourier components. These methods would not be able to create such reflectance curves.

My proposed method would be able to create steep reflectance changes, but only as a last resort when flatter-sloped curves are not able to match the target tristimulus values. The other advantage of the minimum slope squared approach is that it can be expressed as a quadratic objective function subject to linear constraints, which is solvable by standard least-square strategies.

Consider this optimization formulation:

$$\text{minimize} \quad \sum_{i=1}^{n-1} (\rho_{i+1} - \rho_i)^2$$
$$\text{s.t.} \quad T\rho = rgb,$$

where $n$ is the number of discrete wavelength bands (36 in this study). This optimization can be solved by solving the system of linear equations arising from the Lagrangian stationary conditions

$$\begin{bmatrix} D & T^\top \\ T & 0 \end{bmatrix} \begin{Bmatrix} \rho \\ \lambda \end{Bmatrix} = \begin{Bmatrix} 0 \\ rgb \end{Bmatrix},$$

where $D$ is a 36x36 tridiagonal matrix

$$D = \begin{bmatrix} 2 & -2 & & & & \\ -2 & 4 & -2 & & & \\ & -2 & 4 & -2 & & \\ & & \ddots & \ddots & \ddots & \\ & & & -2 & 4 & -2 \\ & & & & -2 & 2 \end{bmatrix}.$$

Since $D$ and $T$ do not depend on $rgb$, the matrix can be inverted ahead of time, instead of each time an sRGB value is processed. Defining

$$B = \begin{bmatrix} D & T^\top \\ T & 0 \end{bmatrix}^{-1} = \begin{bmatrix} B_{11} & B_{12} \\ B_{21} & B_{22} \end{bmatrix},$$

we have

$$\begin{Bmatrix} \rho \\ \lambda \end{Bmatrix} = \begin{bmatrix} B_{11} & B_{12} \\ B_{21} & B_{22} \end{bmatrix} \begin{Bmatrix} 0 \\ rgb \end{Bmatrix}$$

or

$$\rho = B_{12}\, rgb,$$

where $B_{12}$ is the upper-right 36x3 portion of the $B$ matrix. Alternatively, the matrix inversion leading to $B_{12}$ can be computed explicitly, yielding

$$B_{12} = (D^\top D + T^\top T - D^\top T^\top (T T^\top)^{-1} T D)^{-1} T^\top$$

This 36×3 $B_{12}$ matrix is shown here.

Since computing $\rho$ is a simple matter of matrix multiplication, the LSS method is just as computationally efficient as the LLS method, and tends to give much better reflectance curves, as I'll demonstrate later.

Here is a Matlab program for the LSS (Least Slope Squared) method. It also works in the open source free alternative to Matlab, called Octave.

**Least Log Slope Squared (LLSS) Method**

It is generally not a good idea to allow reflectance curves with negative values. Not only is this physically meaningless, but it also can cause problems down the road when the reflectance curves are used in other computations. For example, when modeling subtractive color mixture[1], it may be necessary to require the reflectance curves to be strictly positive.

One way to modify the LSS method to keep reflectances positive is to operate the algorithm in the space of the logarithm of reflectance values, $z = \ln(\rho)$. I call this the **Least Log Slope Squared (LLSS) method**:

$$\begin{aligned} &\text{minimize} \quad \sum_{i=1}^{35} (z_{i+1} - z_i)^2 \\ &\text{s.t.} \quad T e^{z} = rgb. \end{aligned}$$

This new optimization is not as easy to solve as the previous one. Nevertheless, the Lagrangian formulation can still be used, giving rise to a system of 39 *nonlinear* equations and 39 unknowns:

$$F = \begin{Bmatrix} Dz - \text{diag}(e^{z})\, T^{\mathsf{T}} \lambda \\ -T\, e^{z} + rgb \end{Bmatrix} = \begin{Bmatrix} 0 \\ 0 \end{Bmatrix},$$

where $D$ is the same 36x36 tridiagonal matrix presented earlier.

Newton's method solves this system of equations with ease, typically in just a few iterations. Forming the Jacobian matrix,

$$J = \left[ \begin{array}{c|c} D - \text{diag}(\text{diag}(e^{z})\, T^{\mathsf{T}} \lambda) & -\text{diag}(e^{z})\, T^{\mathsf{T}} \\ \hline -T\, \text{diag}(e^{z}) & 0 \end{array} \right],$$

the change in the variables with each Newton iteration is found by solving the linear system

$$J \begin{Bmatrix} \Delta z \\ \Delta \lambda \end{Bmatrix} = -F.$$

Here is a Matlab program for the LLSS (Least Log Slope Squared) method. I added a check for the special case of sRGB = (0,0,0), which simply returns $\rho = (0.0001, 0.0001, \ldots, 0.0001)$. This is necessary since the log formulation is not able to create a reflectance of exactly zero (nor is that desirable in some applications). It can come very close to zero as $z$ approaches $-\infty$, but it is numerically better to handle this one case specially. I chose the value of 0.0001 because it is the largest power of ten that translates back to an integer sRGB triplet of (0,0,0).

The LLSS method requires substantially more computational effort than the previous two methods. Each iteration of Newton's method requires the solution of 39 linear equations in 39 unknowns.

This Matlab program has also been tested in Octave and was found to work fine.

**Iterative Least Log Slope Squared (ILLSS) Method**

The LLSS method above can return reflectance curves with values >1. Although this is physically meaningful phenomenon (fluorescent objects can exhibit this), it may be desirable in some applications to have the entire reflectance curve between 0 and 1. It dawned on me that I might be able to modify the Lagrangian formulation to cap the reflectance values at 1. The main obstacle to doing this is that the use of inequality constraints in the Lagrangian approach greatly complicates the solution process, requiring the solution of the "KKT conditions," and in particular, the myriad "complementary slackness" conditions. If only there were some way to know which reflectance values need to be constrained at 1, then these could be treated by a set of equality constraints and no KKT solution would be necessary.

That led me to investigate the nature of the LLSS reflectance curves with values >1. I ran the LLSS routine on every value of sRGB by intervals of five, that is, sRGB = (0,0,0), (0,0,5), (0,0,10), …, (255,255,250), (255,255,255). In every one of those 140,608 cases, the algorithm found a solution in less than a dozen or so iterations (usually just a handful), and 38,445 (27.3%) of them had reflectance values >1.

Of the 38,445 solutions with values >1, 36,032 of them had a single contiguous region of reflectance values >1. The remaining 2,413 had two regions, always located at both ends of the visible spectrum. Since the distribution of values >1 is so well defined, I started thinking of an algorithm that would iteratively force the reflectance to 1. It would start by running the LLSS method. If any of the reflectance values ended up >1, I would add a single equality constraint forcing the reflectance at the maximum of each contiguous >1 region to equal 1, solve that optimization, then force the adjacent values that were still >1 to 1, optimize again, and repeat until all values were $\leq$1.

That was getting to be an algorithmic headache to implement, so I tried a simpler approach, as follows. First, run the LLSS method. If any reflectance values end up >1, constrain ALL of them to equal 1, and re-solve the optimization. This will usually cause some more values adjacent to the old contiguous regions to become >1, so constrain them in addition to the previous ones. Re-solve the optimization. Repeat the last two steps until all values are $\leq$1. Here is an animation of this process, which I call the **Iterative Least Log Slope Squared** (ILLSS) process, applied to sRGB = (75, 255, 255):

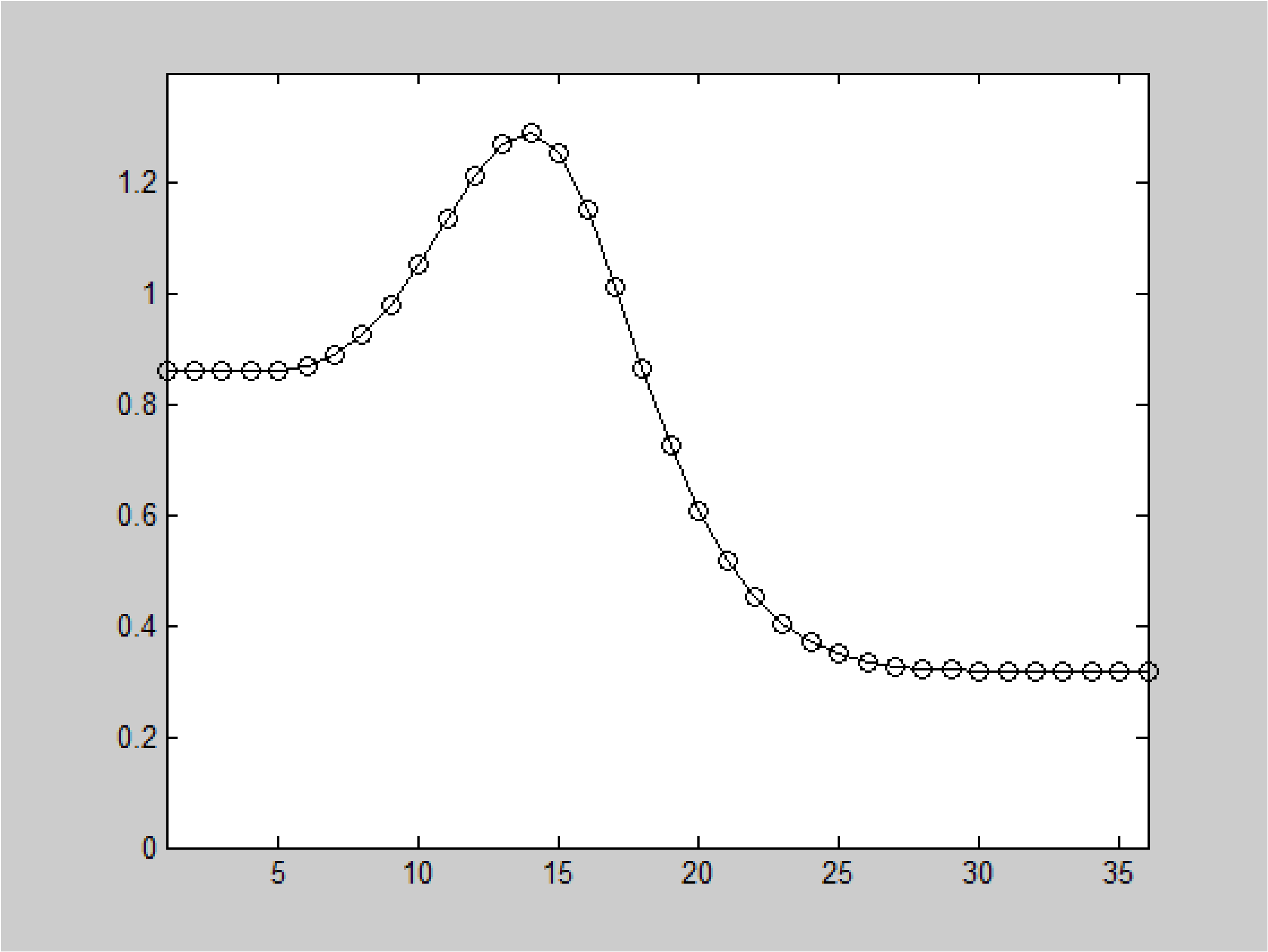

Animation of the ILLSS process (click image link to animate).

To express the ILLSS algorithm mathematically, let's begin with the LLSS optimization statement and add the additional equality constraints:

$$\begin{aligned} &\text{minimize} \quad \sum_{i=1}^{35} (z_{i+1} - z_i)^2 \\ &\text{s.t.} \quad Te^{z} = rgb \\ &\qquad z_k = 0, \quad k \in \text{FixedSet}. \end{aligned}$$

"FixedSet" is the set of reflectance indices that are constrained to equal 1, or equivalently, the set of $z$ indices constrained to equal zero (since $z = \ln(\rho)$ ). Initially, FixedSet is set to be the empty set. Each time the optimization is repeated, the $z$ values $\geq$0 have their indices added to this set.

We can define a matrix that summarizes the fixed set, for example:

$$K = \begin{bmatrix} 0 & 0 & 1 & 0 & 0 & 0 & \dots & 0 \\ 0 & 0 & 0 & 0 & 1 & 0 & \dots & 0 \end{bmatrix}_{2x36}.$$

This example indicates that there are two reflectance values being constrained (because $K$ has two rows), and the third and fifth reflectance values are the particular ones being constrained.

The Lagrangian formulation now has additional Lagrange multipliers, called $\mu$, one for each of the constrained reflectances. The system of nonlinear equations produced by finding a stationary point of the Lagrangian (setting partial derivatives of the Lagrangian with respect to each set of variables ($z$, $\lambda$, and $\mu$) equal to zero) is

$$F = \begin{Bmatrix} Dz - \mathrm{diag}(e^{z})\, T^{\mathsf{T}}\lambda + K^{\mathsf{T}}\mu \\ -T\, e^{z} + rgb \\ Kz \end{Bmatrix} = \begin{Bmatrix} 0 \\ 0 \\ 0 \end{Bmatrix},$$

where $D$ is the same 36x36 tridiagonal matrix presented earlier. As before, we solve this nonlinear system with Newton's method. Forming the Jacobian matrix,

$$J = \left[\begin{array}{c|c|c} D - \mathrm{diag}(\mathrm{diag}(e^{z})\, T^{\mathsf{T}}\lambda) & -\mathrm{diag}(e^{z})\, T^{\mathsf{T}} & K^{\mathsf{T}} \\ \hline -T\, \mathrm{diag}(e^{z}) & 0 & 0 \\ \hline K & 0 & 0 \end{array}\right],$$

the change in the variables with each Newton iteration is found by solving the linear system

$$J \begin{Bmatrix} \Delta z \\ \Delta\lambda \\ \Delta\mu \end{Bmatrix} = -F.$$

Here is a Matlab program that performs the ILLSS (Iterative Least Log Slope Squared) optimization. I included a check for the two special cases of $rgb$ = (0,0,0) or (255,255,255), which simply return $\rho$ = (0.0001, 0.0001, …, 0.0001) or (1, 1, …, 1). The additional special case of (255,255,255) is needed because numerical issues arise if the $K$ matrix grows to 36x36, as it would in that second special case. This program works in Octave as well.

**Iterative Least Slope Squared (ILSS) Method**

For completeness, I thought it would be a good idea to add one more algorithm. Recall the ILLSS method modifies the LLSS method to cap reflectances >1. Similarly, the ILSS method will modify the LSS method to cap values both >1 and <0. The ILSS may reduce computational effort in comparison to the ILLSS method since the inner loop of the ILLSS method requires an iterative Newton's method solution, whereas there would be no inner loop needed with the ILSS method; it is simply the solution of a linear system of equations. Here is the ILSS formulation:

$$\begin{aligned} \text{minimize} \quad & \sum_{i=1}^{35} (\rho_{i+1} - \rho_i)^2 \\ \text{s.t.} \quad & T\rho = rgb \\ & \rho_{k_1} = 1, \quad k_1 \in \mathrm{FixedSet}_1 \\ & \rho_{k_0} = \rho_{min}, \quad k_0 \in \mathrm{FixedSet}_0. \end{aligned}$$

$\mathsf{FixedSet}_1$ is the set of reflectance indices that are constrained to equal 1, and $\mathsf{FixedSet}_0$ is the set of indices that are constrained to equal $\rho_{min}$ (the smallest allowable reflectance, typically 0.00001). Initially, both fixed sets are the empty set. Each time the optimization is repeated, the $\rho$ values $\geq 1$ have their indices added to $\mathsf{FixedSet}_1$ and those $\leq \rho_{min}$ have their indices added to $\mathsf{FixedSet}_0$.

We define two matrices that summarize the fixed sets, for example:

$$K_1 = \begin{bmatrix} 0 & 0 & 1 & 0 & 0 & 0 & \dots & 0 \\ 0 & 0 & 0 & 0 & 1 & 0 & \dots & 0 \end{bmatrix}_{2x36}$$

$$K_0 = \begin{bmatrix} 1 & 0 & 0 & 0 & 0 & 0 & \dots & 0 \\ 0 & 0 & 0 & 0 & 0 & 1 & \dots & 0 \\ 0 & 0 & 0 & 1 & 0 & 0 & \dots & 0 \end{bmatrix}_{3x36} .$$

This example indicates that there are two reflectance values being constrained to equal 1 (because $K_1$ has two rows), and the third and fifth reflectance values are the particular ones being constrained. There are three reflectance values being constrained to $\rho_{min}$ (because $K_0$ has three rows), and the first, sixth, and fourth are the particular ones being constrained. The order in which the rows appear in these matrices is not important.

At each iteration of the ILSS method, this linear system is solved:

$$\begin{bmatrix} D & T^{\mathsf{T}} & K_1^{\mathsf{T}} & K_0^{\mathsf{T}} \\ T & 0 & 0 & 0 \\ K_1 & 0 & 0 & 0 \\ K_0 & 0 & 0 & 0 \end{bmatrix} \begin{Bmatrix} \rho \\ \lambda \\ \mu \\ \nu \end{Bmatrix} = \begin{Bmatrix} 0 \\ rgb \\ 1 \\ \rho_{min} \end{Bmatrix},$$

where $\mu$ and $\nu$ are Lagrange multipliers associated with the $K_1$ and $K_0$ sets, respectively.

This system can be solved for $\rho$, yielding the expression

$$\rho = R - B_{11} K^{\mathsf{T}} \left[ K B_{11} K^{\mathsf{T}} \right]^{-1} \left( KR - \begin{Bmatrix} 1 \\ \rho_{min} \end{Bmatrix} \right),$$

where

$$B = \begin{bmatrix} D & T^{\mathsf{T}} \\ T & 0 \end{bmatrix}^{-1}, \quad R = B_{12}\, rgb, \quad \text{and} \quad K = \begin{bmatrix} K_1 \\ K_0 \end{bmatrix}.$$

$B_{11}$ and $B_{12}$ are the upper-left 36x36 and upper-right 36x3 parts of $B$, respectively. They can be computed ahead of time. Note that only an $m$x$m$ matrix needs to be inverted at each iteration, where $m$ is the number of $\rho$ values being held fixed, typically zero or a small number. When $m$ is zero, the ILSS method simplifies to the LSS method.

Here is a Matlab program that performs the ILSS (Iterative Least Slope Squared) method. It also works in Octave.

## Comparison of Methods

***(Update 1/9/20: The newly-developed LHTSS method presented below is a better choice than ILLSS when requiring the reflectance curves to fall strictly between 0 and 1. It matches the Munsell colors better and takes about 60% of the computational effort of ILLSS. See the 6/4/2019 update below.)***

I ran each of the five algorithms with every sRGB value (by fives) and have summarized the results in the table below. The total number of runs for each solution was 140,608.

| Name | Execution Time (sec) | Max $\rho$ | Min $\rho$ | Num $\rho$ >1 | Num $\rho$ <0 | Max Iter. | Mean Iter. | Computational Effort |
|---|---|---|---|---|---|---|---|---|
| **LLS, Linear Least Squares** | 2.7 | 1.36 | -0.28 | 26,317 | 50,337 | n/a | n/a | Matrix mult only |
| **LSS, Least Slope Squared** | 2.7 | 1.17 | -0.17 | 9,316 | 48,164 | n/a | n/a | Matrix mult only |
| **ILSS, Iterative Least Slope Squared** | 25.7 | 1 | 0 | 0 | 0 | 5 | 1.49 | Mult soln of $m$ linear eqns** |
| **LLSS, Least Log Slope Squared** | 322. | 3.09 | 0 | 38,445 | 0 | 16 | 6.77 | Mult soln of 39 linear eqns |
| **ILLSS, Iterative Least Log Slope Squared** | 525. | 1 | 0 | 0 | 0 | 5* | 1.41* | Mult soln of (39+ $m$) linear eqns** |

* This is outer loop iteration count. The inner loop has iteration count similar to previous line.
** The quantity $m$ is the number of reflectance values that end up being constrained at either 1 or 0.

### Explanation of Columns

- Execution Time: Real-time duration to compute 140,608 reflectance curves of all sRGB values (in intervals of five), on a relatively slow Thinkpad X61 tablet. Relative times are more important than absolute times.
- Max $\rho$: The maximum reflectance value of all computed curves.
- Min $\rho$: The minimum reflectance value of all computed curves. Zero is used to represent some small specified lower bound, typically 0.0001.
- Num $\rho$>1: The number of reflectance curves with maxima above 1.
- Num $\rho$<0: The number of reflectance curves with minima below 0.
- Max Iter.: The largest number of iterations required for any reflectance curve (for iterative methods).
- Mean Iter.: The mean value of all iteration counts (for iterative methods).
- Computational Effort: Comments on the type of computation required for the method.

Clearly, the methods that don't need to solve linear systems of equations are much faster. The reflectance curves they create, however, may not be very realistic.

## Comparison of Reflectance Curves

In this section, I compare the computed reflectance curves against two large sets of measured reflectance data. My goal is to assess how “realistic” the computed reflectance curves are in comparison to reflectances measured from real colored objects.

## Munsell Color Samples

The first set comes from the Munsell Book of Colors, specifically the 2007 glossy version. The Munsell system organizes colors by hue, chroma (like saturation), and value (like lightness). In 2012, Paul Centore measured 1485 different Munsell samples and published the results here.[7] I computed the sRGB values of each sample, and used those quantities to compute reflectance curves for each of the five methods described above. Of the 1485 samples, 189 of them are outside the sRGB gamut (have values <0 or >255) and were not examined in this study, leaving 1296 in-gamut samples. An Excel file of the 1485 reflectance curves and sRGB values is available here.[8]

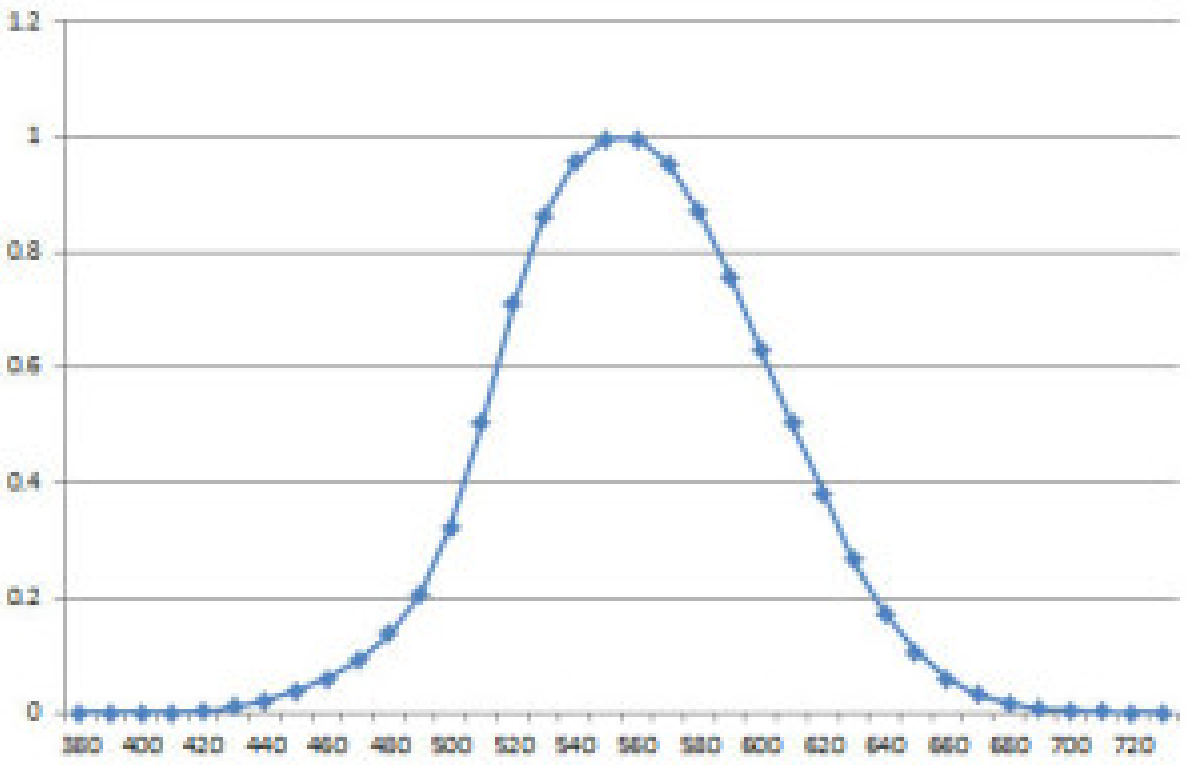

The luminous efficiency curve, which describes the relative sensitivity of the eye to spectral light across the visible spectrum.

When comparing reflectance curves, some parts of the curve are more import than others. As we approach both ends of the visible spectrum, adjacent to the UV and IR ranges, the human eye’s sensitivity to these wavelengths rapidly decreases. This phenomenon is described by the “luminous efficiency” curve[9] shown in the adjacent figure. As I was computing the reflectance curves, I noticed that there are often large discrepancies between the computed and measured reflectance curves at the ends of the visible spectrum. These large differences have relatively little impact on color perception, and would also have little consequence on operations performed on the computed reflectance curves, such as when modeling subtractive color mixture. Consequently, I decided to downplay these end differences when comparing the curves.

To assess how well each computed reflectance matches the measured reflectance curve, I first subtracted one from the other and took absolute values of the difference. Then I multiplied the differences by the luminous efficiency curve. Finally, I summed all of the values to give a single measure of how well the reflectance curves match, or a “reflectance match measure” ($RMM$):

$$RMM = \sum_{i=380}^{730} lum_i \left|\left(\rho_i^{measured} - \rho_i^{computed}\right)\right|.$$

Lower values of $RMM$ represent a better match, with zero being a perfect match of the two reflectance curves. Keep in mind that regardless of the value of $RMM$, the sRGB triplets of the computed and measured curves always match exactly, as would the perceived color evoked by the two reflectance curves.

I examined the maximum and mean values of $RMM$ for each of the five methods when applied to all 1296 Munsell samples:

| **Name** | **Max $RMM$** | **Mean $RMM$** |
|---|---|---|
| LLS, Linear Least Squares | 2.46 | 0.88 |
| LSS, Least Slope Squared | 1.11 | 0.17 |
| ILSS, Iterative Least Slope Squared | 1.04 | 0.16 |
| LLSS, Least Log Slope Squared | 0.92 | 0.15 |
| ILLSS, Iterative Least Log Slope Squared | 0.86 | 0.15 |

The Linear Least Squares method is clearly much worse than the others. The following figures compare the last four methods. Each gray square represents one Munsell sample, and the shade of gray indicates the corresponding $RMM$ value. The shade of gray is linearly mapped so that it is white for $RMM$ = 0 and black for $RMM$ = 1.11, the maximum value $RMM$ for all four methods.

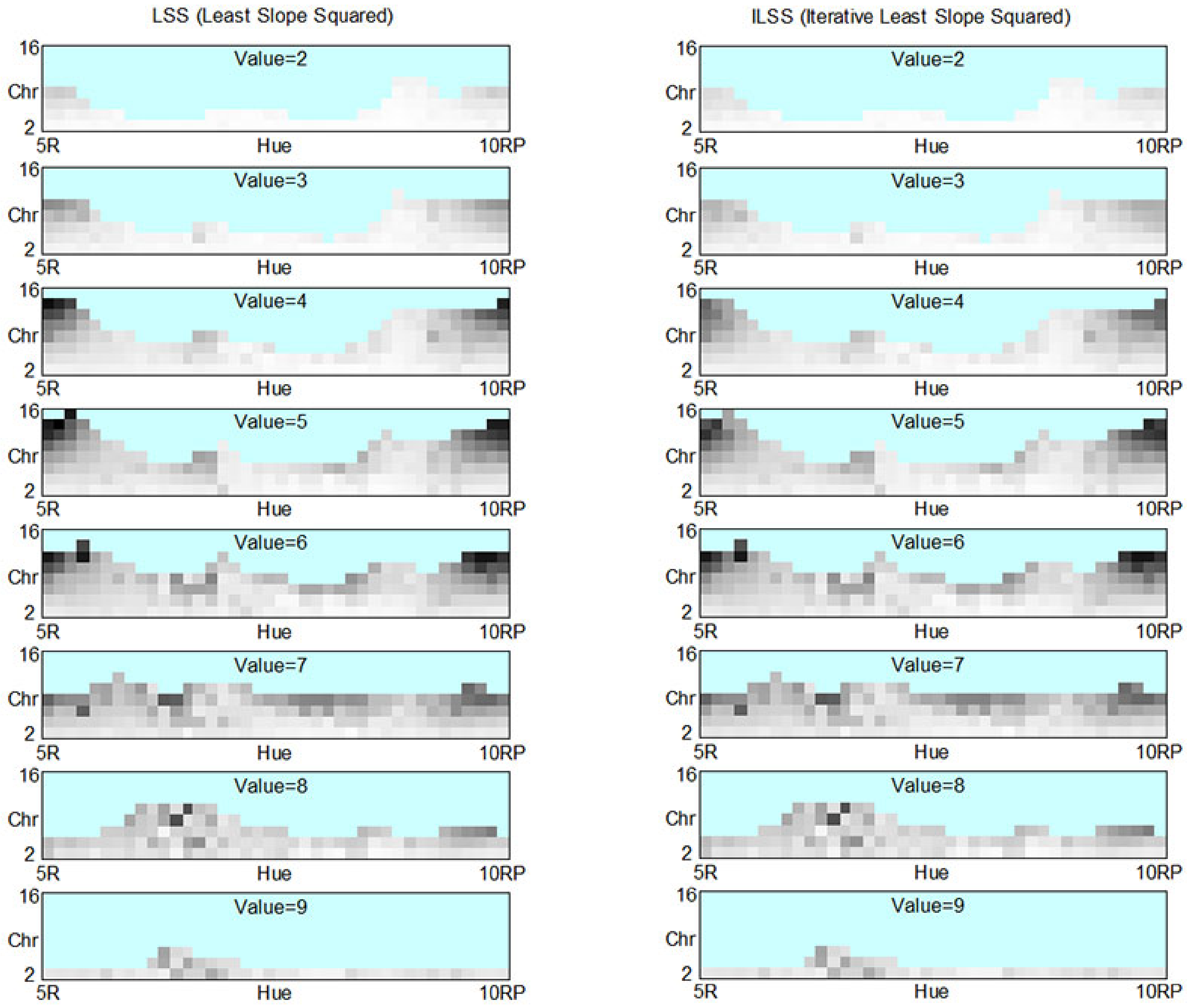

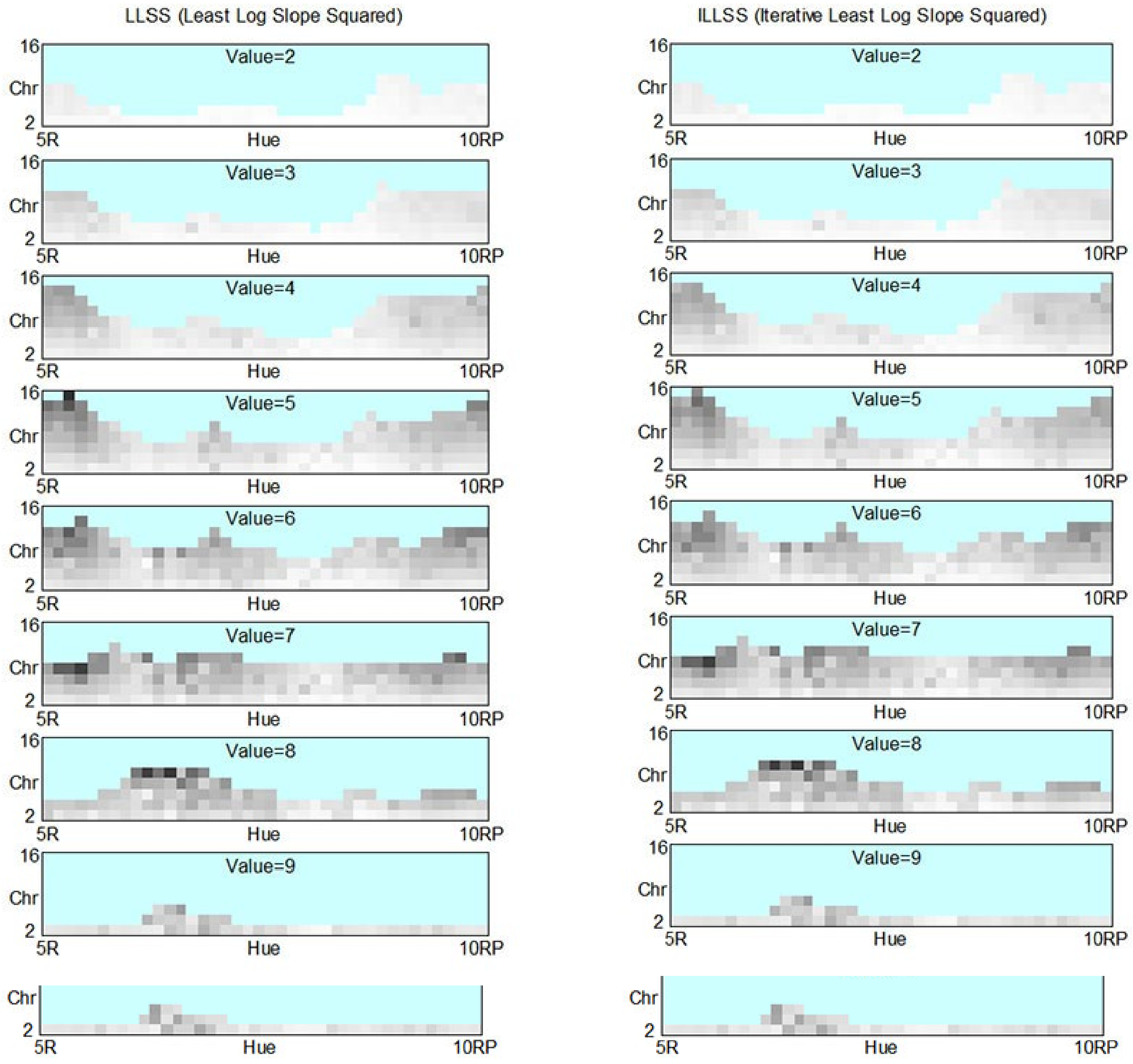

Overall, the log versions of the algorithms do a little better. The worst matches with non-log-based LSS/ILSS take place with the highly chromatic reds and purples, and sometimes with the bright yellows. With log-based LLSS/ILLSS, the worst matches tend to be in the bright yellows, and sometimes in the chromatic reds and oranges. In both pairs, the iterative version clips the reflectance curves between 0 and 1, and usually reduces the mismatch somewhat, but not by very much.

Here are some examples of LSS and ILSS with high $RMM$s:

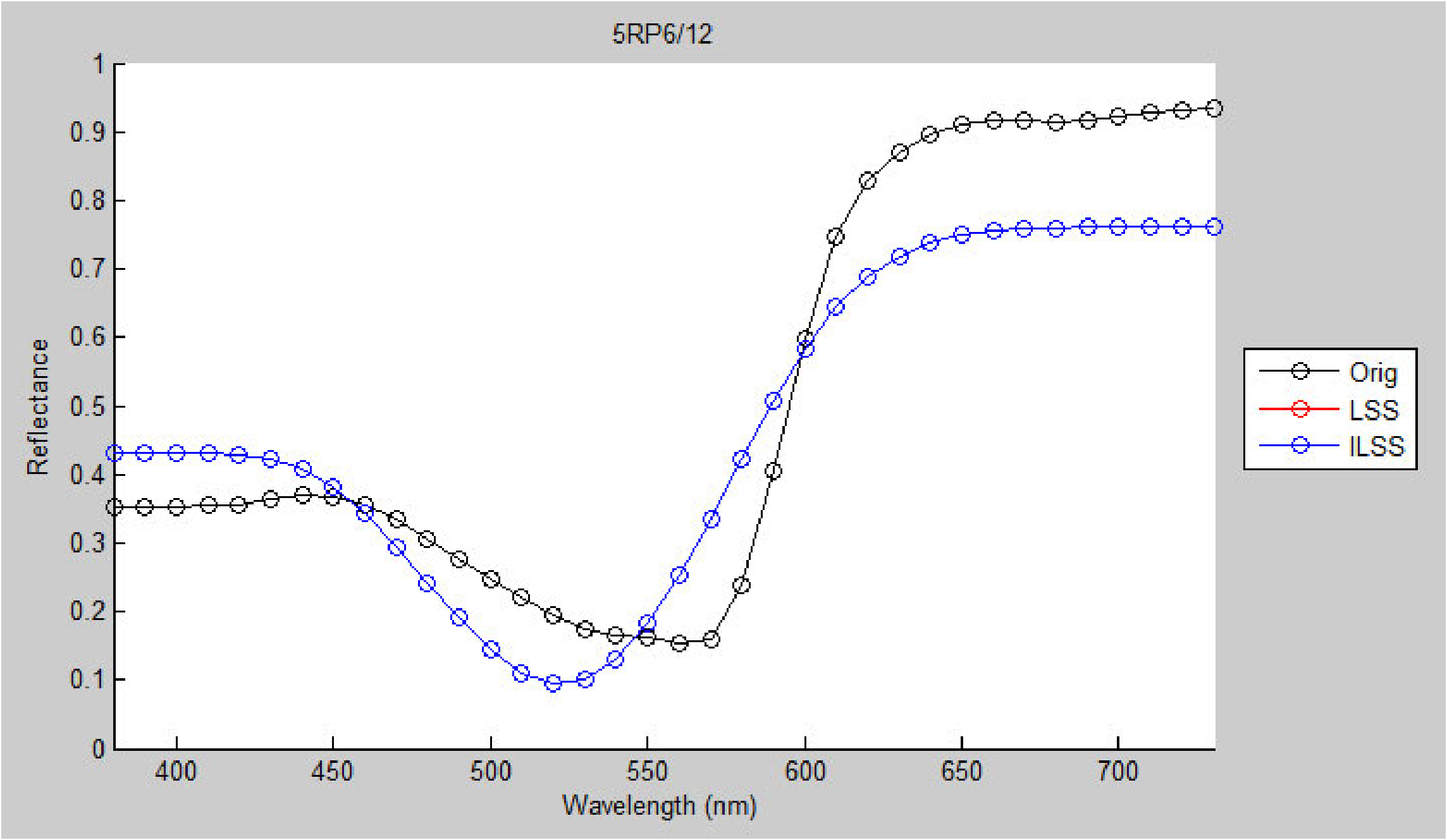

This is typical of the mismatch with chromatic reds and purples. This one (Hue=5RP, Value=6, Chroma=12) has an $RMM$ of 1.04. The curve for LSS is not visible because it is directly behind the ILSS curve. There was no clipping needed, so the two curves are the same.

One weakness of the LSS method is that it tends to drop into the negative region for chromatic red colors. In these cases, the clipping provided by the ILSS method greatly improves the match. The following example has an LSS $RMM$ of 1.01 and an improved ILSS $RMM$ of 0.40:

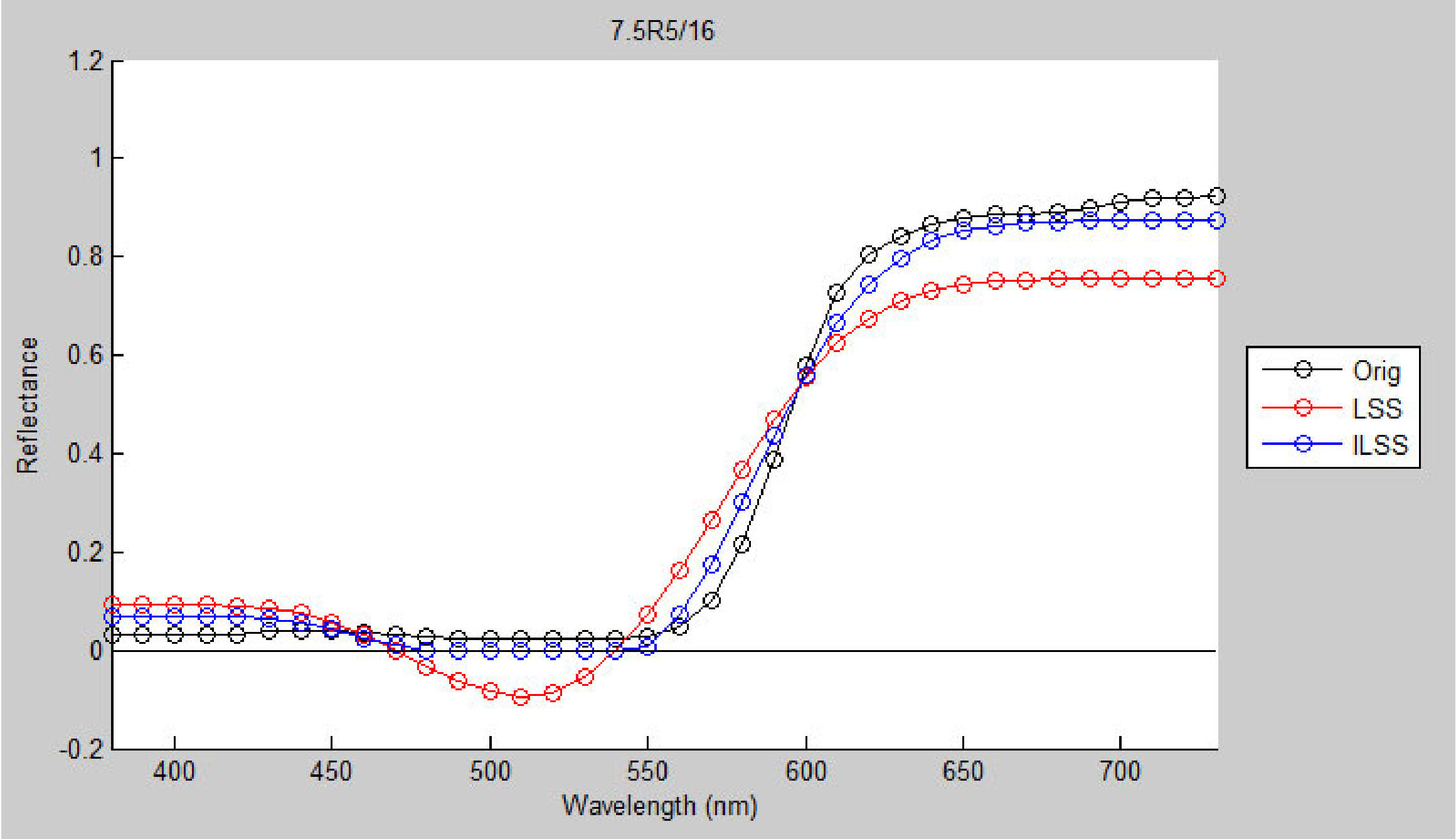

The following is a typical mismatch for LSS/ILSS in the yellow region:

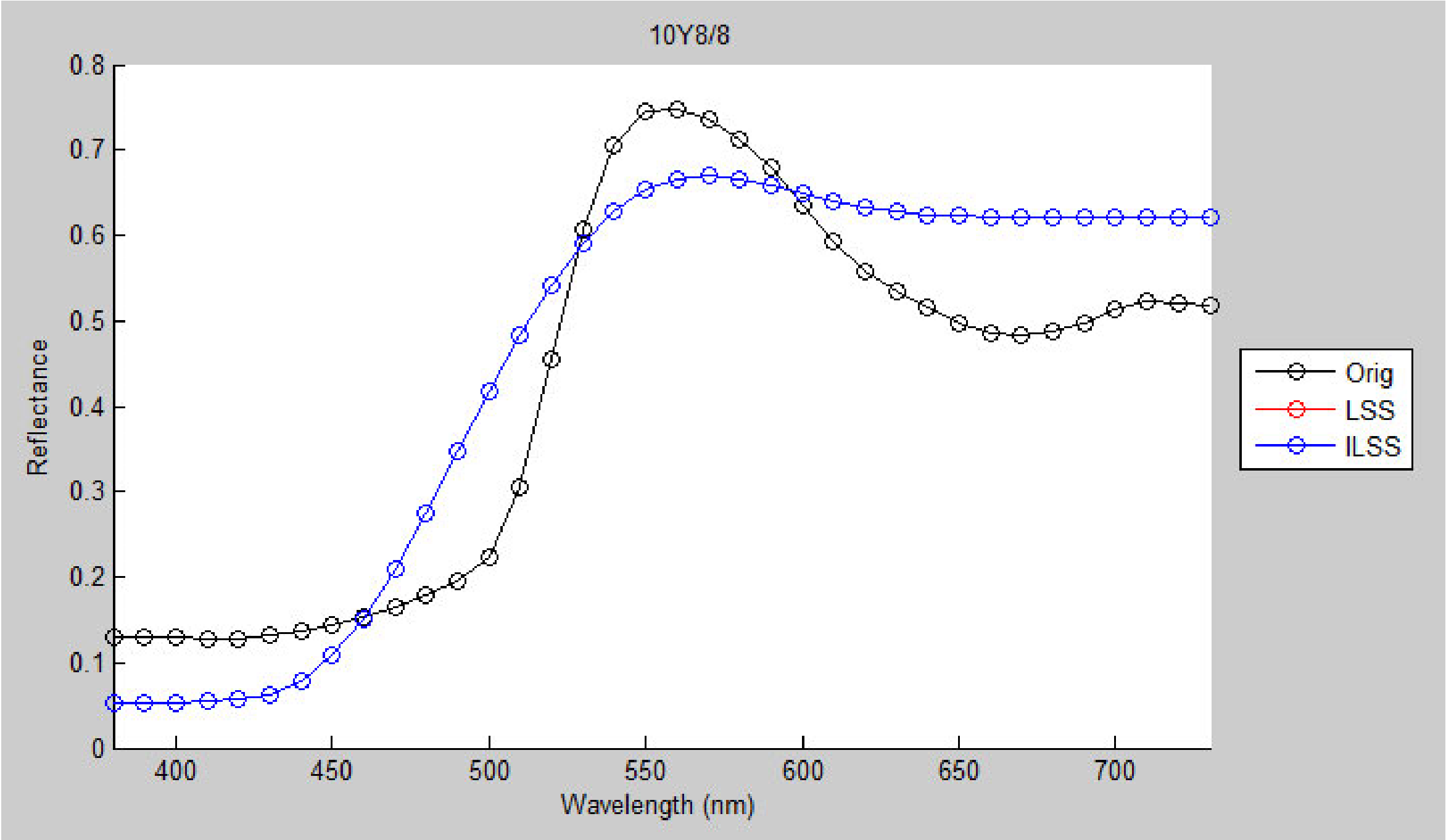

Generally, when LSS/ILSS has a bad match, it is because it oscillates too little.

Here are some examples of log-based LLSS/ILLSS with high $RMM$s:

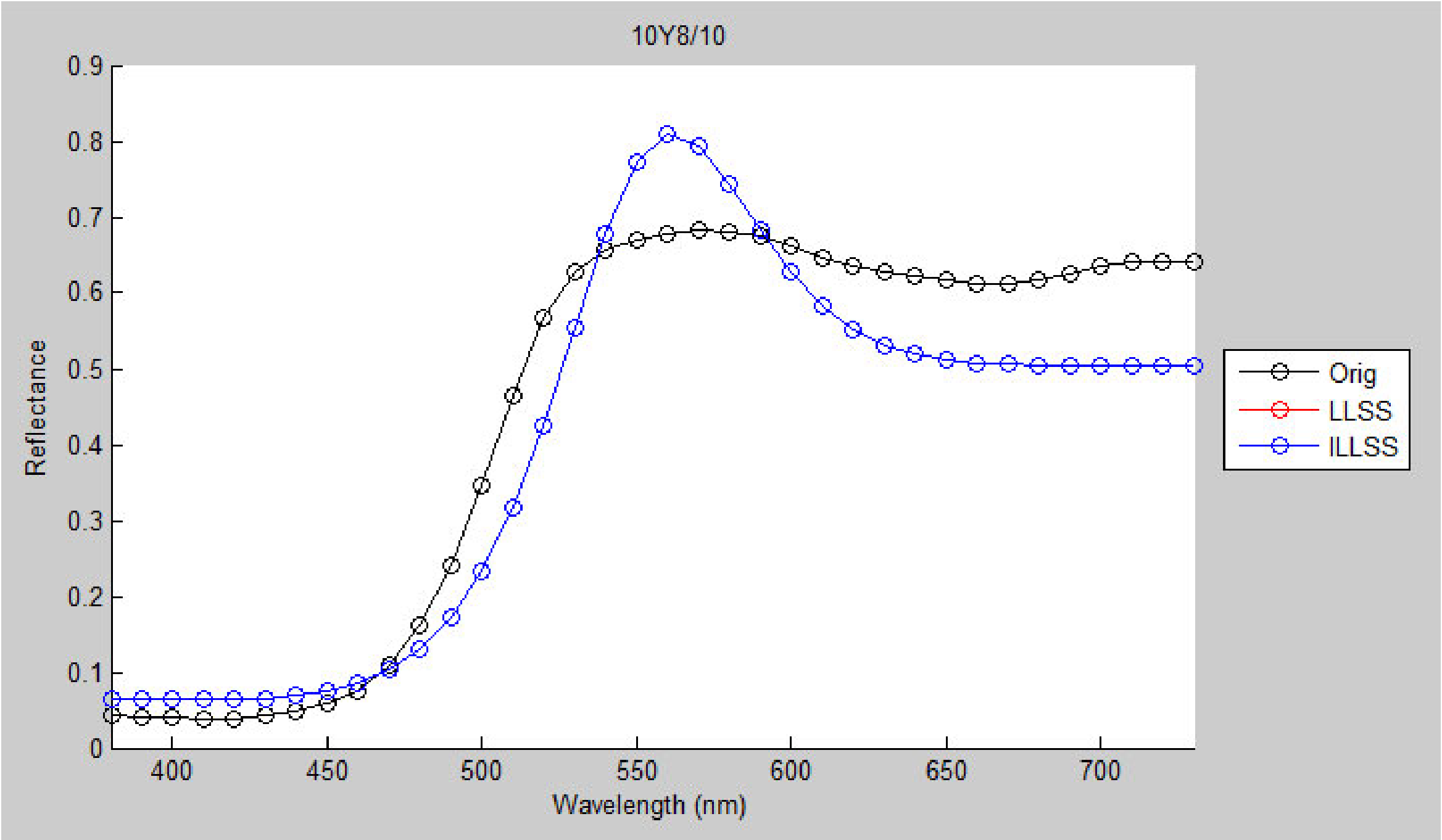

This previous example has an $RMM$ of 0.88, mainly because it takes advantage of the wavelengths of light that give rise to a yellow sensation, in contrast to how the pigments in this sample get the same yellow sensation from a broader mix of wavelengths.

On bright red and orange Munsell samples, the LLSS/ILLSS curves tend to shoot high on the red side:

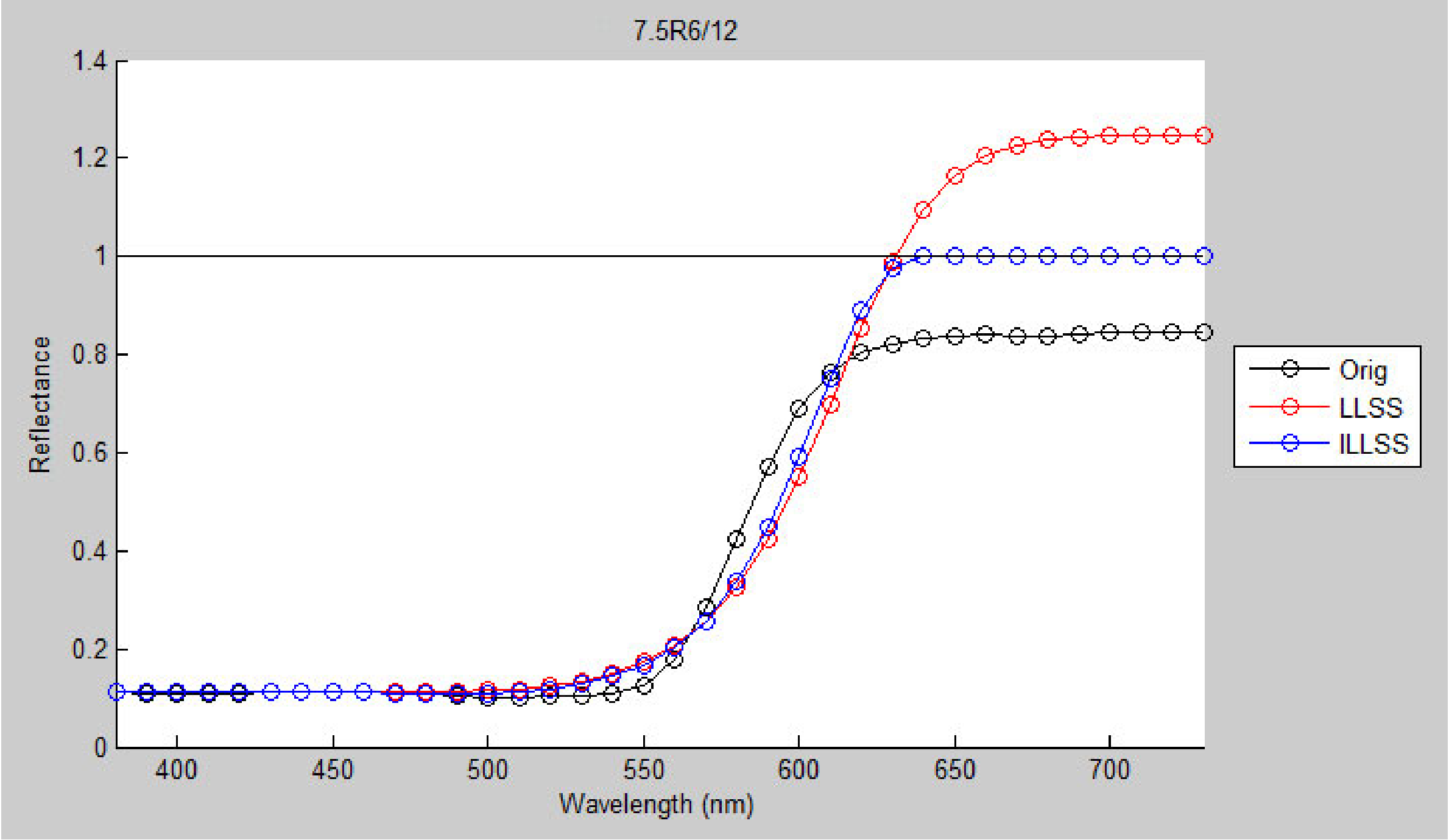

Even though there is considerable clipping with the ILLSS version, most of it takes place in the long wavelengths, which does not help the $RMM$ score as much. This is evident in how little ILLSS needs to adjust the mid-wavelength range to make up for the huge difference in the long wavelengths, while maintaining the same sRGB values.

In summary, the non-log versions (LSS/ILSS) tend to undershoot peaks and the log versions (LLSS/ILLSS) tend to overshoot them. Overall, the log versions give a somewhat better match, but at the expense of considerably more computation.

**Commercial Paints and Pigments**

The second large dataset of reflectance curves comes from Zsolt Kovacs-Vajna's RS2color webpage[10]. He has a database of reflectance curves for many sets of commercial paints and pigments, which can be obtained by emailing a request to him. They are grouped into the following 31 families:

| | |
|---|---|
| 1. apaFerrario_PenColor | 17. MunsellGlossy5G |
| 2. Chroma_AtelierInteractive | 18. MunsellGlossy5P |
| 3. ETAC_EFX500b | 19. MunsellGlossy5R |
| 4. ETAC_EFX500t | 20. MunsellGlossy5Y |
| 5. GamblinConservationColors | 21. MunsellGlossyN |
| 6. Golden_HB | 22. pigments |
| 7. Golden_OpenAcrylics | 23. supports |
| 8. GretagMacbethMini | 24. Talens_Ecoline |
| 9. Holbein_Aeroflash | 25. Talens_Rembrandt |
| 10. Holbein_DuoP | 26. Talens_VanGoghH2Oil |
| 11. KremerHistorical | 27. whites |
| 12. Liquitex_HB | 28. WinsorNewton_ArtAcryl |
| 13. Maimeri_Acqua | 29. WinsorNewton_Artisan |
| 14. Maimeri_Brera | 30. WinsorNewton_Finity |
| 15. Maimeri_Polycolor | 31. WinsorNewtonHandbook |
| 16. MunsellGlossy5B | |

In total, there are 1493 different samples in these 31 groups. I discarded the 275 of them that were out of the sRGB gamut, and computed reflectance curves from the sRGB values of the remaining 1218 samples. I again computed $RMM$ values to compare the computed curves to the actual measured curves. The following animated GIFs show the $RMM$ values (color coded according to the colormap on the right) plotted in sRGB space.

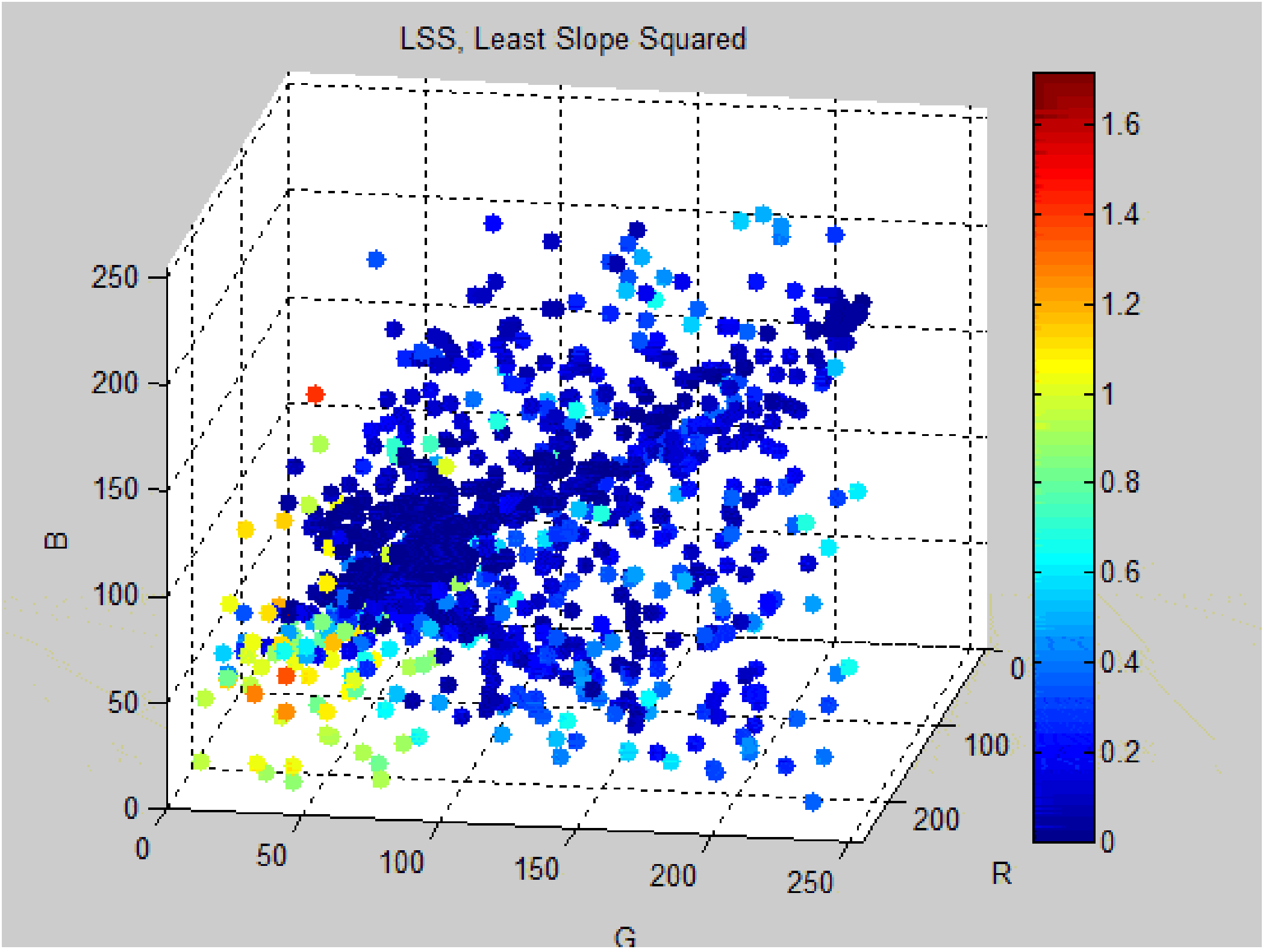

Above, $RMM$ values (by color) for the LSS method plotted in sRGB space.
(Click on link to view animated GIF.)

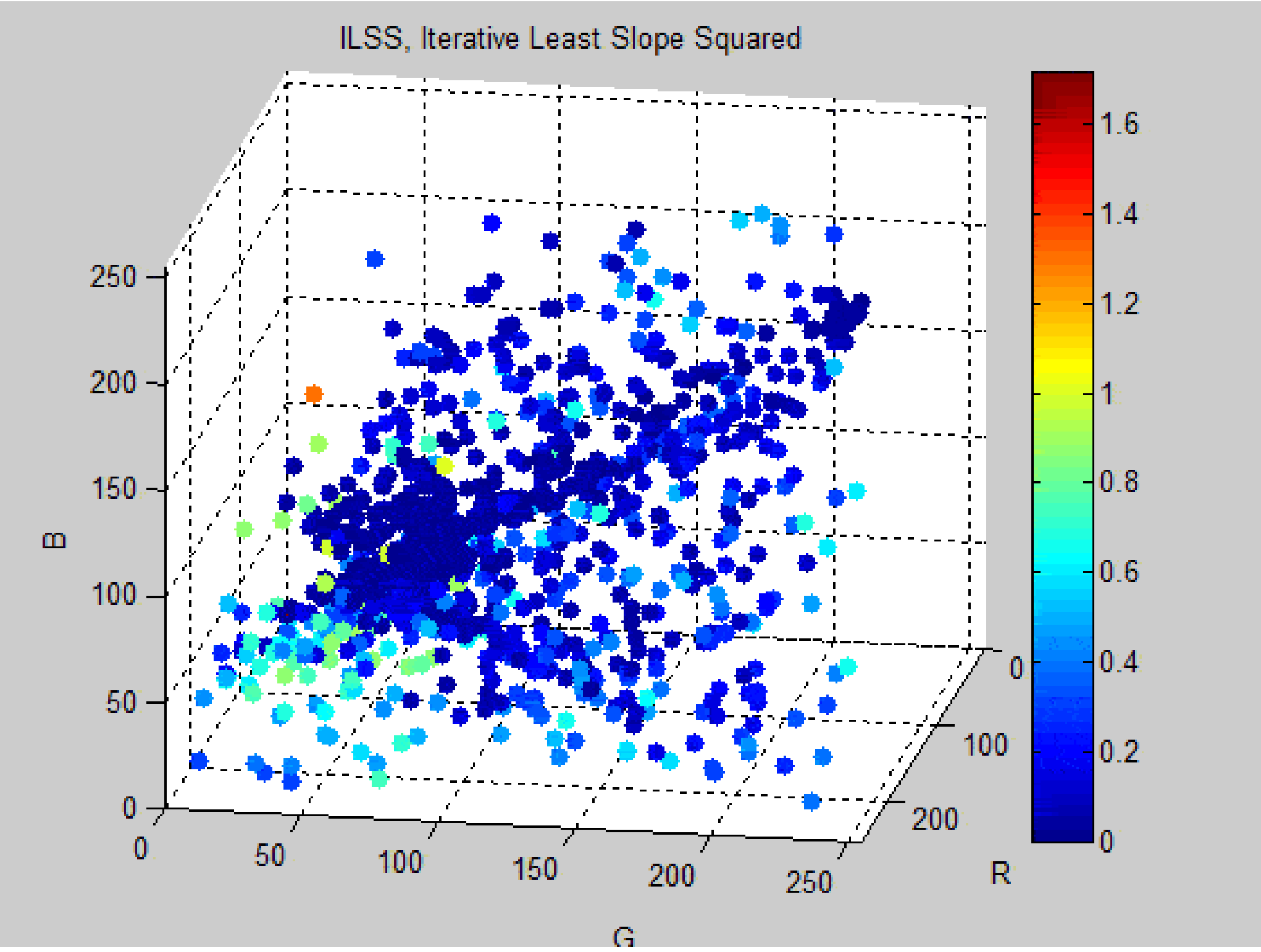

Above, $RMM$ values (by color) for the ILSS method plotted in sRGB space. (Click for GIF)

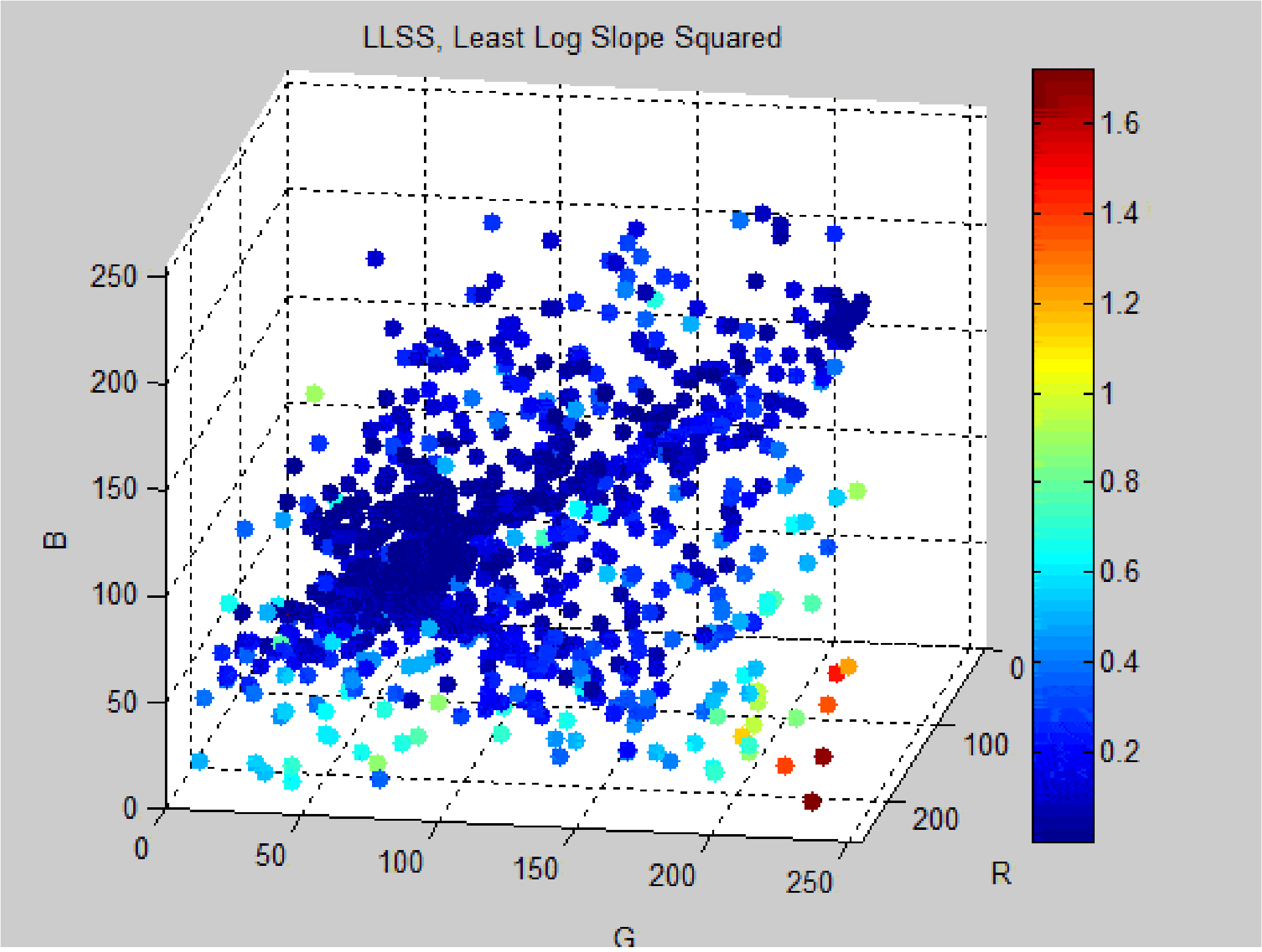

Above, $RMM$ values (by color) for the LLSS method plotted in sRGB space. (Click for GIF)

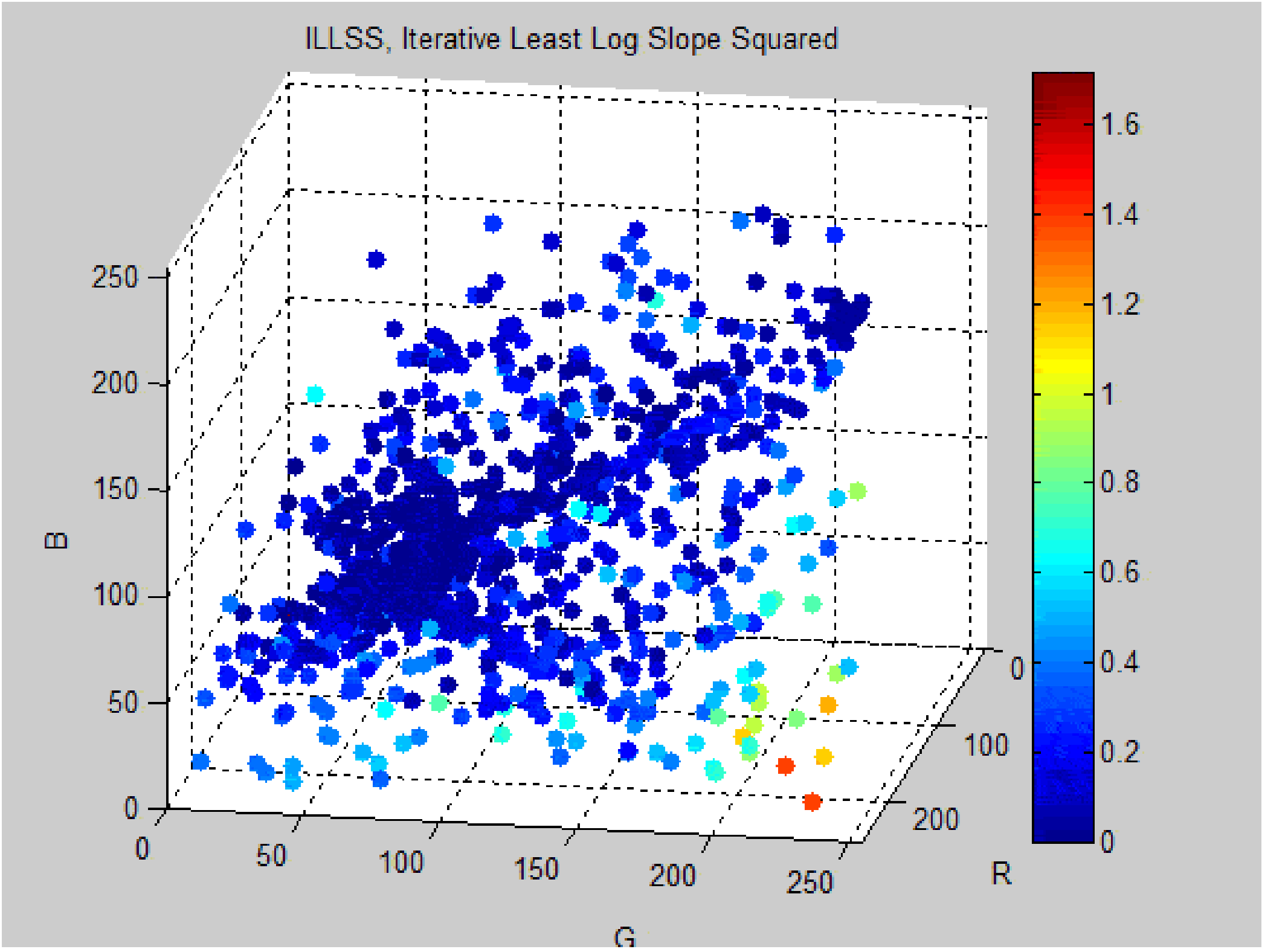

Above, $RMM$ values (by color) for the ILLSS method plotted in sRGB space. (Click for GIF)

It is apparent from these GIFs that the non-log-based methods have a fairly large region of mismatch in the red region (large R, small G and B). The log-based versions have a much smaller region of mismatch in the yellow region (large R and G, small B). The iterative version of both methods improve the matches in both cases. The relative sizes of the mismatch regions can be seen more clearly in a projection onto the R-G plane:

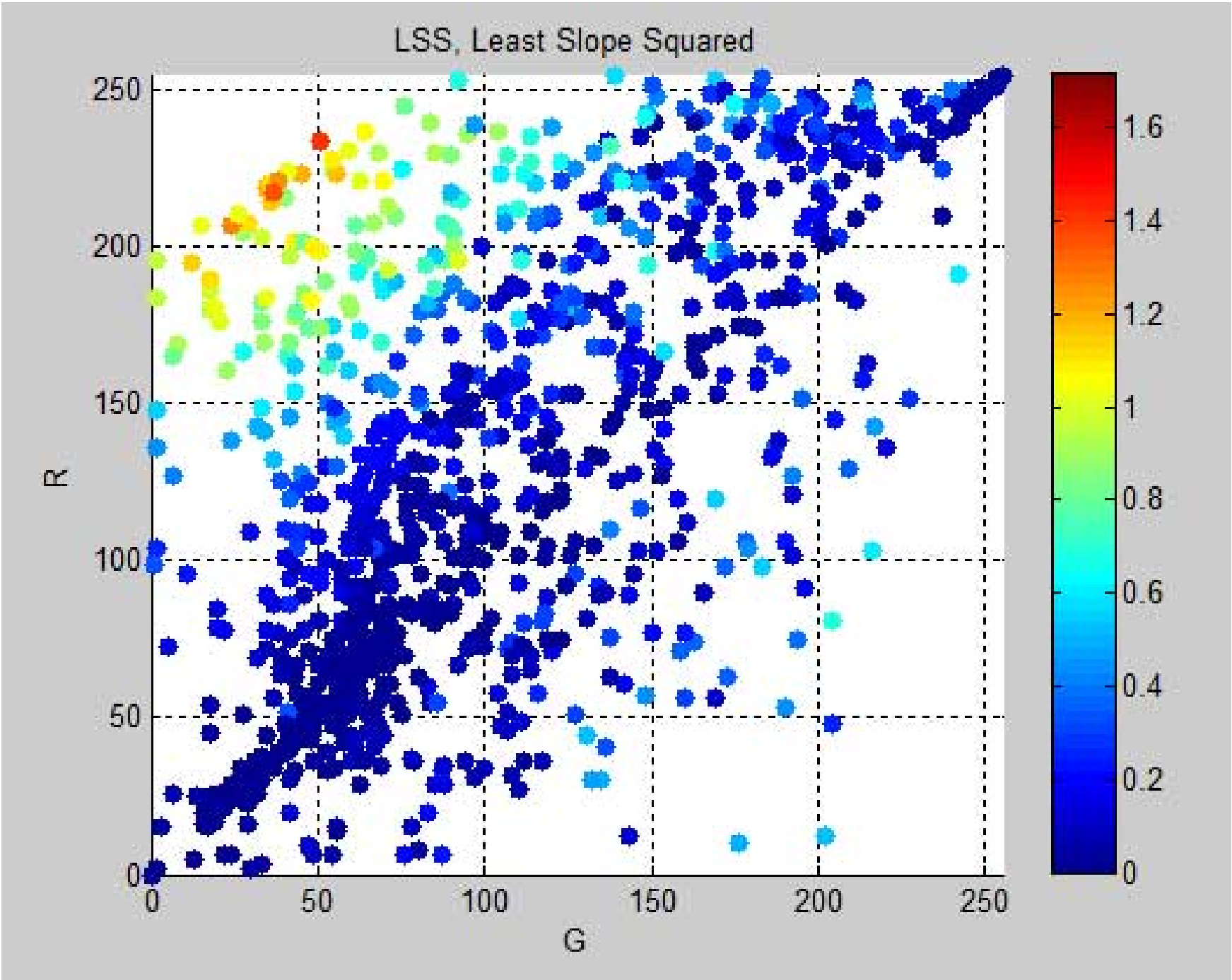

Above, $RMM$ values (by color) for the LSS method projected onto the sRGB R-G plane.

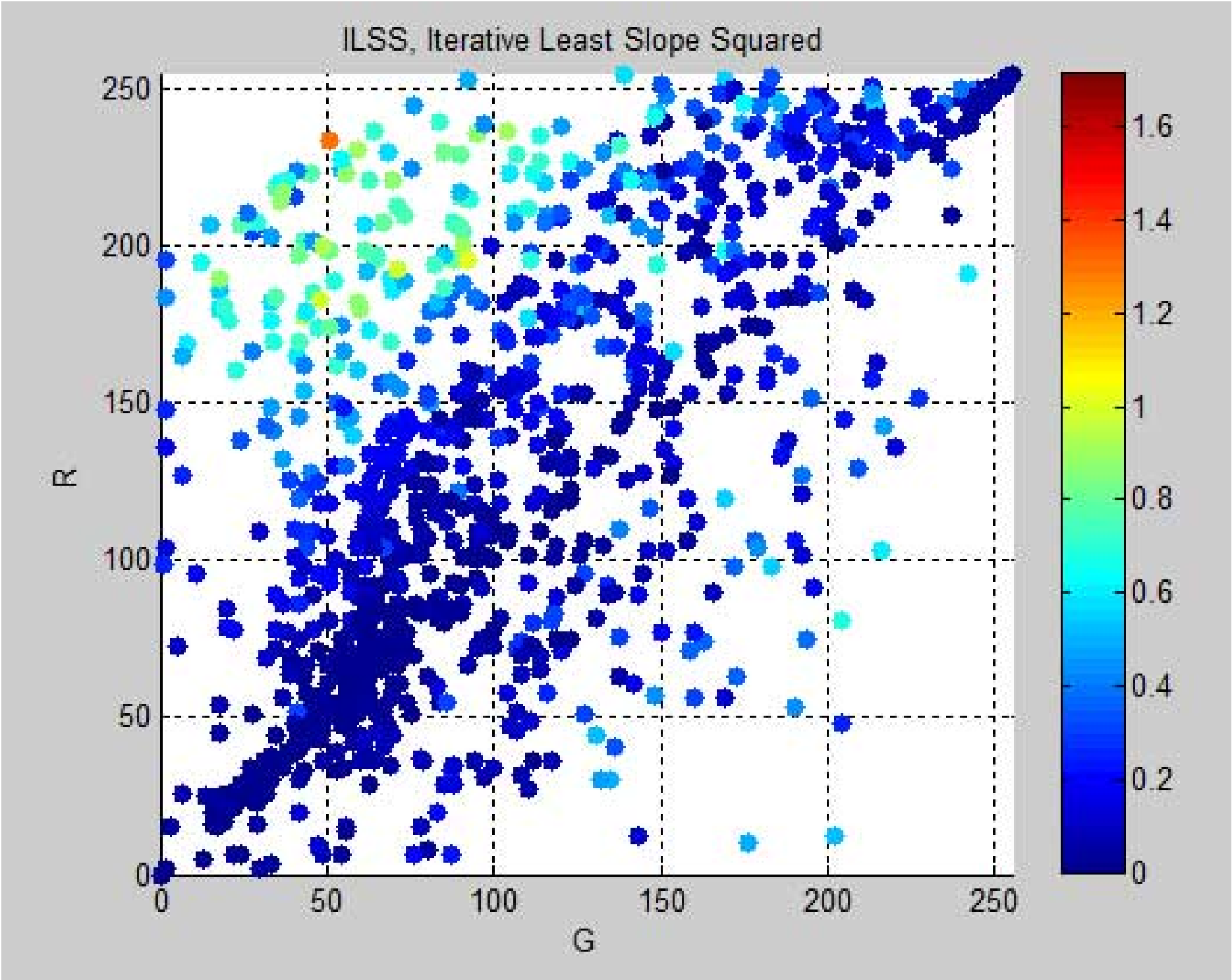

Above, $RMM$ values (by color) for the ILSS method projected onto the sRGB R-G plane.

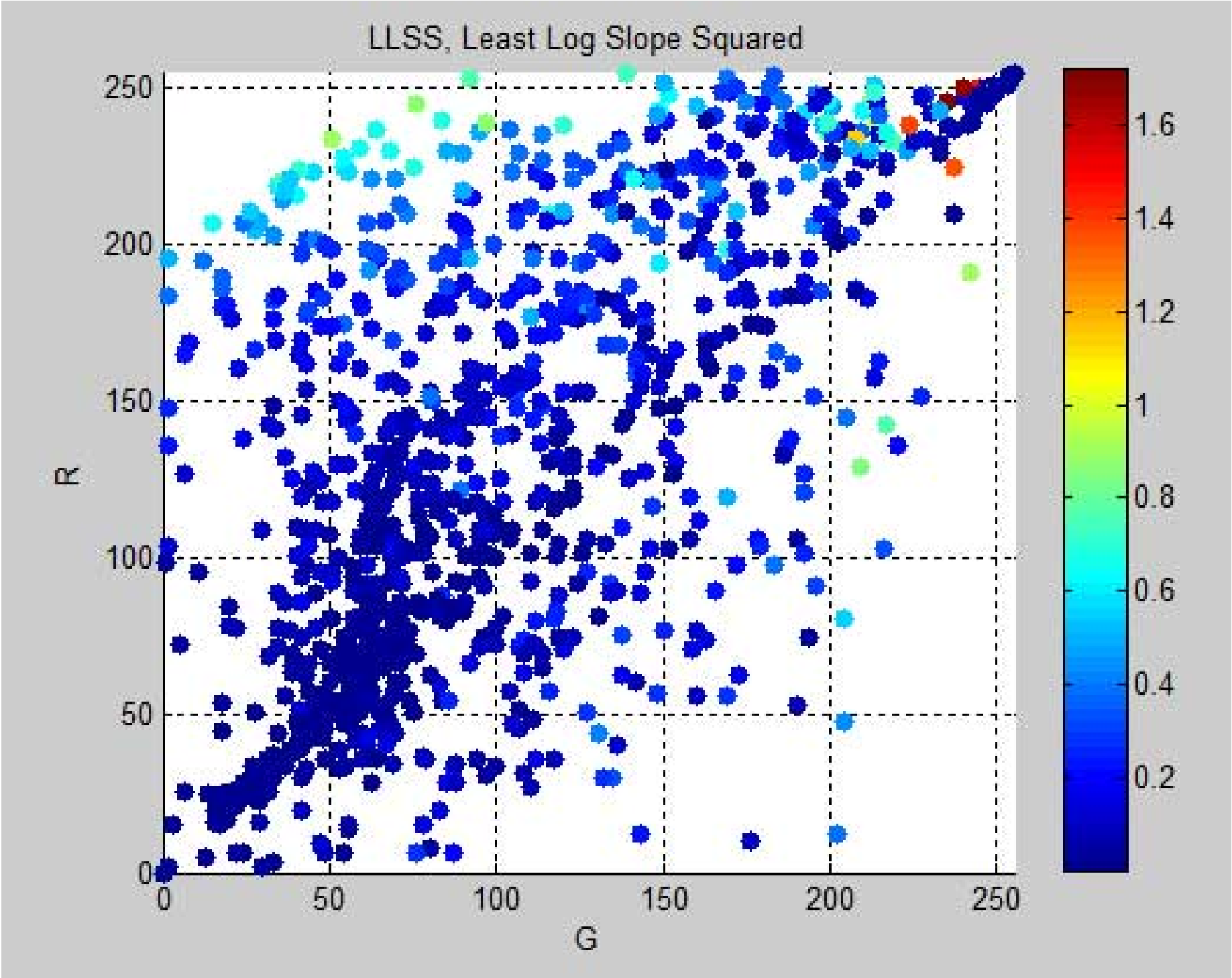

Above, $RMM$ values (by color) for the LLSS method projected onto the sRGB R-G plane.

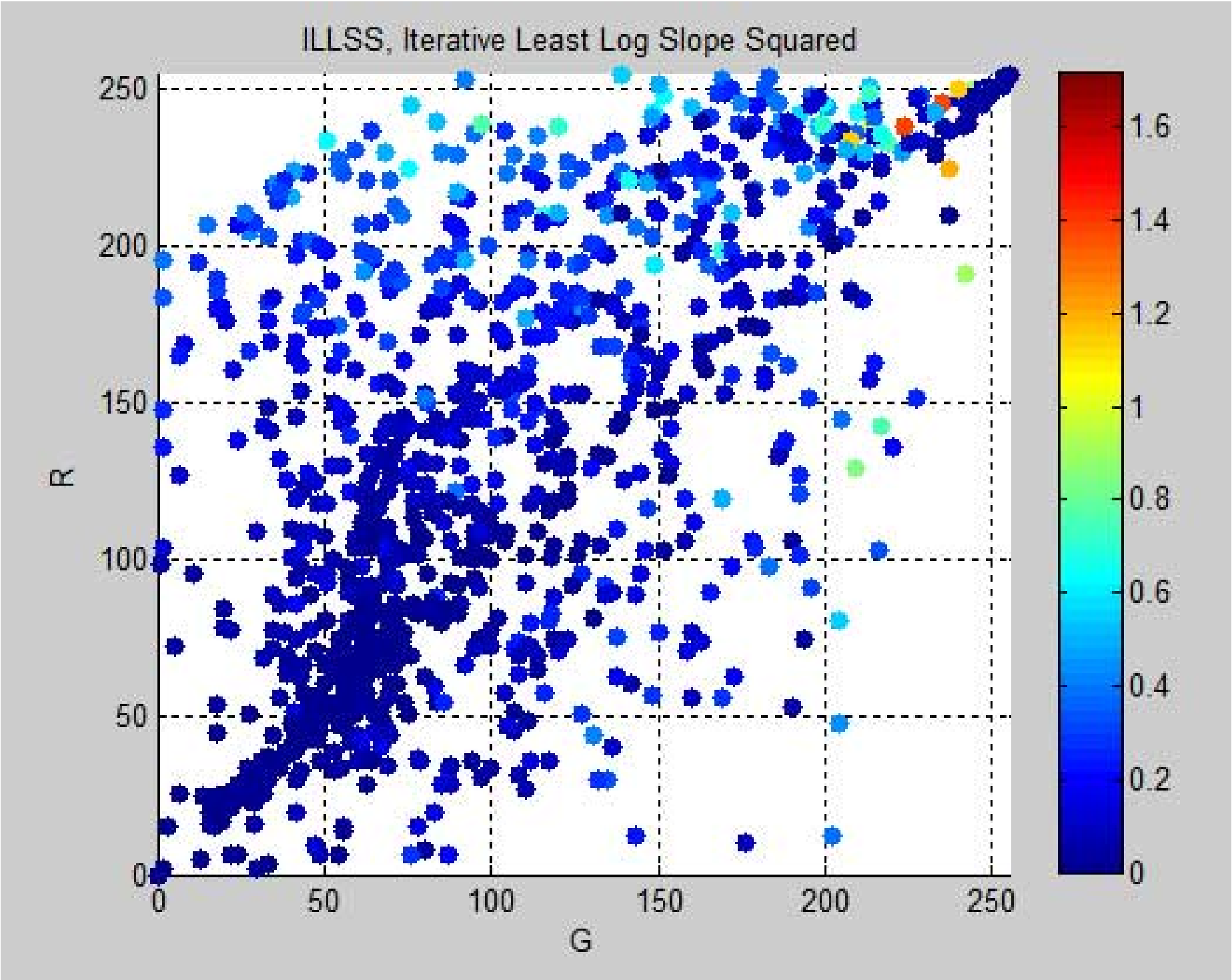

Above, $RMM$ values (by color) for the ILLSS method projected onto the sRGB R-G plane.

## Conclusions

***(Update 1/9/20: The newly-developed LHTSS method presented below is a better choice than ILLSS when requiring the reflectance curves to fall strictly between 0 and 1. It matches the Munsell colors better and takes about 60% of the computational effort of ILLSS. See the 6/4/2019 update below.)***

In summary, the method that best matches paint and pigment colors found commercially and in nature is the ILLSS (Iterative Least Log Slope Squared) method. It suffers, however, from very large computational requirements. If efficiency is more important, then the ILSS (Iterative Least Slope Squared) method is the preferred one. A third alternative that is midway between the match quality and computing effort is LLSS (Least Log Slope Squared), but be aware that some reflectance curves can end up with values >1. I would not recommend the LSS (Least Slope Squared) method, despite its spectacular computational efficiency, because it can give reflectance curves with physically meaningless negative values. The following table summarizes the three suggested methods:

| Algorithm Name | Computational Effort | Comments | Link to Matlab/Octave Code |
|---|---|---|---|
| **ILSS (Iterative Least Slope Squared)** | Relatively little. | Very fast, but tends to undershoot reflectance curve peaks, especially for bright red and purple colors. Always returns reflectance values in the range 0-1. | link |
| **LLSS (Least Log Slope Squared)** | About 12 times that of ILSS. | Better quality matches overall, but tends to overshoot peaks in the yellow region. Some reflectance values can be >1, especially for bright red colors. | link |
| **ILLSS (Iterative Least Log Slope Squared)** | About 20 times that of ILSS. | Best quality matches. Tends to overshoot peaks in the yellow region. Always returns reflectance values in the range 0-1. | link |

---

***Update 6/4/2019:*** *I've just developed a new method that gives even better natural-reflectance-matching results than the ILLSS method above, with considerably less computational effort required (comparable to that of LLSS). Here it is:*

**Least Hyperbolic Tangent Slope Squared (LHTSS) Method**

Recall that the Least Log Slope Squared (LLSS) method generates reflectance curves that are strictly positive, without requiring explicit bounds in the optimization statement. The lower bound is handled implicitly in the logarithmic transformation. This section presents a newly developed approach to generate reflectances that are strictly within the 0-1 range, also not needing explicit bounds. Why is this useful? The methods that use explicit bounds presented above (ILSS and ILLSS) tend to have abrupt discontinuities in slope when a bounding constraint is engaged. This makes the reflectance curves seem somewhat unnatural. The LHTSS method described in this section makes reflectance curves that gently approach the upper and/or lower bounds, and match the reflectance curves of commercial paints and pigments even better than any of the previous methods above.

The **Least Hyperbolic Tangent Slope Squared (LHTSS) method** uses a transformation defined by

$$\rho = \frac{\tanh(z) + 1}{2}$$

to keep all reflectance values between 0 and 1. Below is a plot of the hyperbolic tangent function:

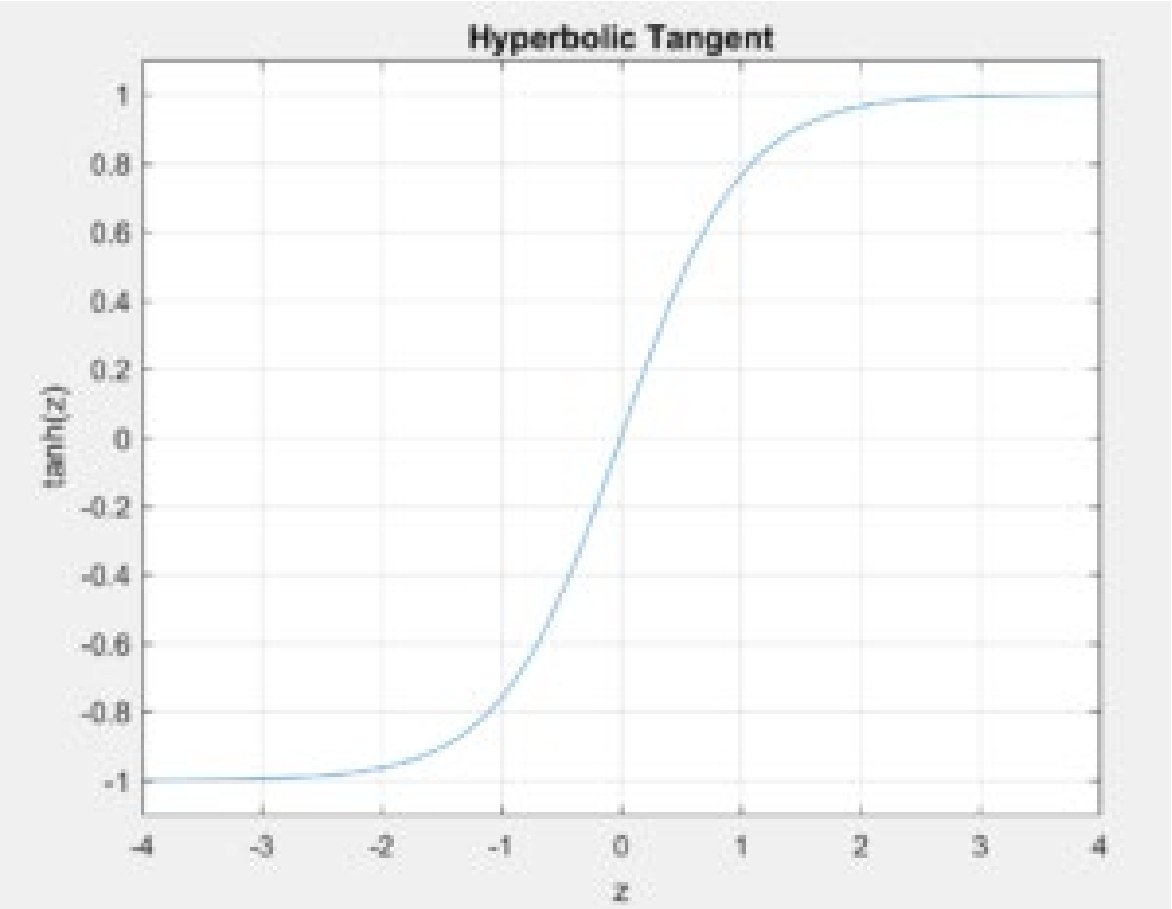

Note how the curve approaches +1/-1 as $z$ grows large in either the positive or negative directions. By adding 1 to tanh and then dividing by 2, the upper and lower bounds shift to 1 and 0. This causes the reflectances to remain strictly between 0 and 1 for any value of $z$, as shown below:

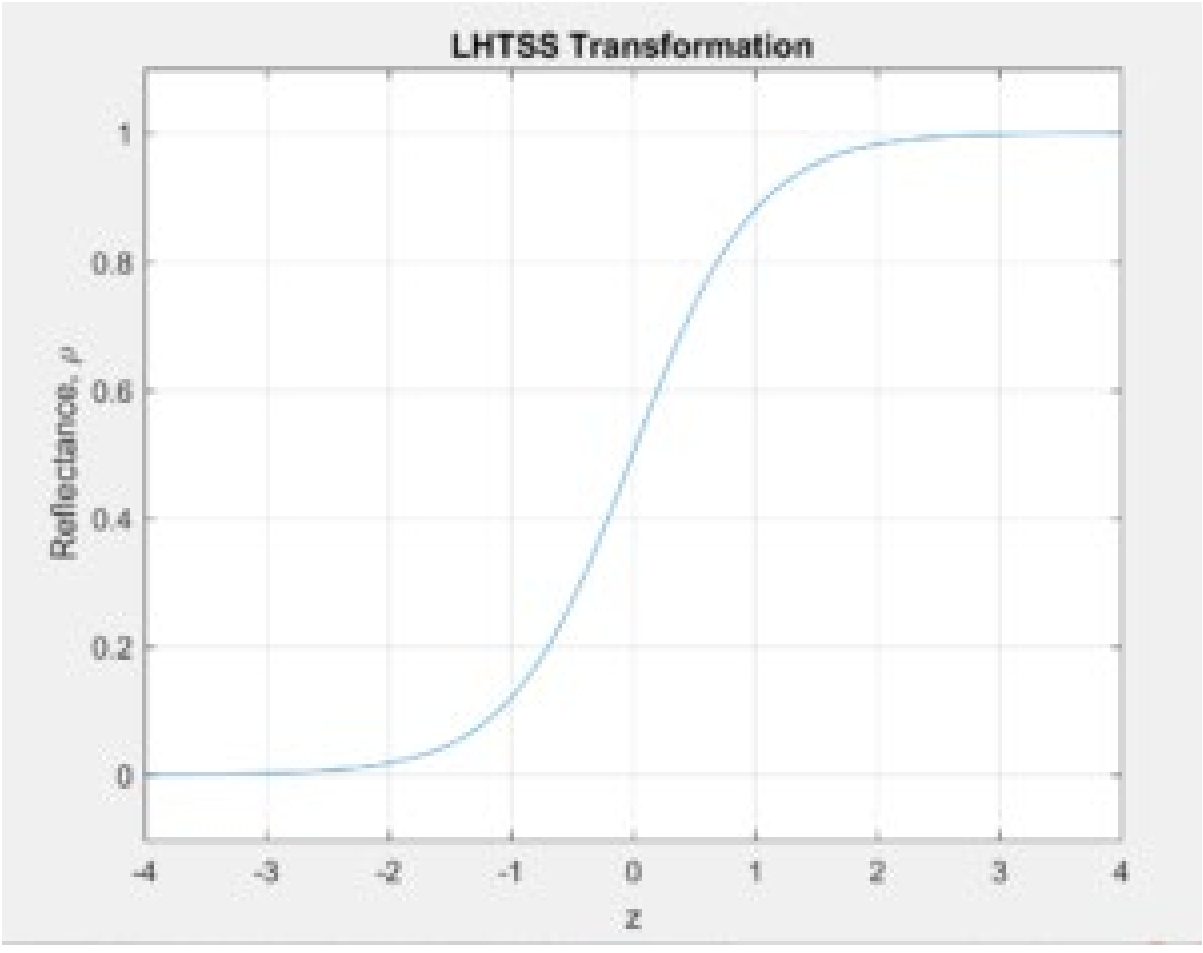

The least slope optimization statement for the Hyperbolic Tangent Slope Squared method is:

$$\text{minimize} \quad \sum_{i=1}^{35} (z_{i+1} - z_i)^2$$
$$\text{s.t.} \quad T\left\{\frac{\tanh(z) + 1}{2}\right\} = rgb.$$

Noting that the first derivative of $\tanh(z) + 1$ is $\text{sech}^2(z)$, the stationary conditions of the Lagrangian function associated with this optimization comprise a system of 39 nonlinear equations and 39 unknowns:

$$F = \begin{Bmatrix} Dz + \text{diag}(\text{sech}^2(z)/2)\ T^\mathsf{T}\lambda \\ T\ \text{sech}^2(z)/2 - rgb \end{Bmatrix} = \begin{Bmatrix} 0 \\ 0 \end{Bmatrix},$$

where $D$ is the 36x36 tridiagonal matrix of finite differencing constants presented earlier.

Newton's method solves this system of equations with ease, typically in a fairly small number iterations. Noting that the first derivative of $\text{sech}^2(z)/2$ is $-\text{sech}^2(z)\tanh(z)$, the Jacobian matrix of first partial derivatives of $F$ is

$$J = \left[\begin{array}{c|c} D - \text{diag}(\text{diag}(\text{sech}^2(z)\tanh(z))\ T^\mathsf{T}\lambda) & \text{diag}(\text{sech}^2(z)/2)\ T^\mathsf{T} \\ \hline T\ \text{diag}(\text{sech}^2(z)/2) & 0 \end{array}\right].$$

The change in the variables with each Newton iteration is found by solving the linear system

$$J \begin{Bmatrix} \Delta z \\ \Delta\lambda \end{Bmatrix} = -F.$$

At each iteration, the values of $z$ and $\lambda$ are updated using $z^{k+1} = z^k + \Delta z$ and $\lambda^{k+1} = \lambda^k + \Delta\lambda$.

Here is a Matlab program for the LHTSS (Least Hyperbolic Tangent Slope Squared) method. It handles the two cases of sRGB = (0,0,0) and (255,255,255) specially, since the reflectance curves associated with them fall outside the strict interval (0->1) imposed by the hyperbolic tangent transform. This Matlab program has also been tested in Octave and was found to work fine.

An example of how this hyperbolic tangent method (LHTSS) compares to the logarithmic method (LLSS) and the regular least slope squared method (LSS), when applied to the Munsell color 7.5R 5/16, is shown in the figure below.

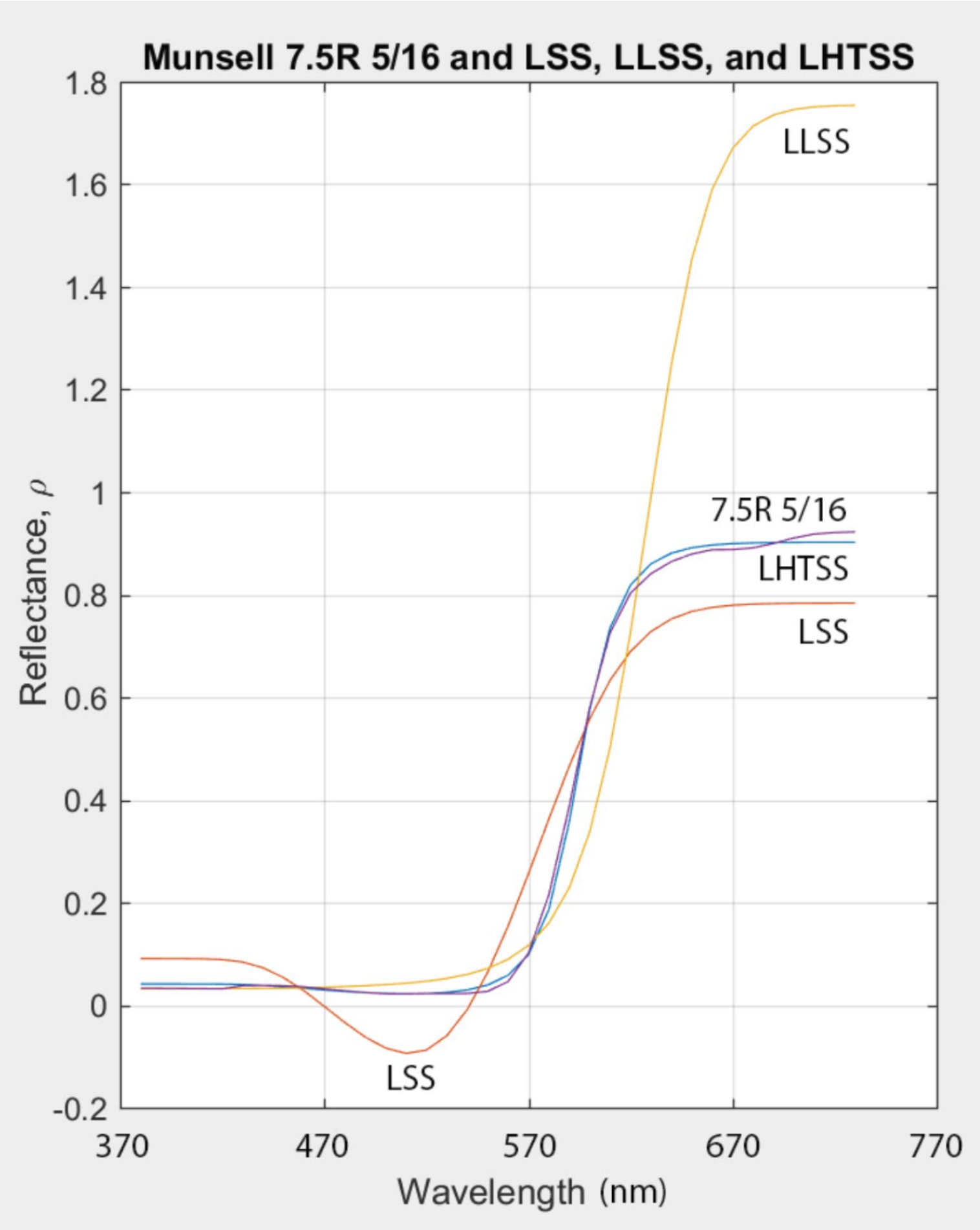

This figure demonstrates an instance where LSS has negative elements, LLSS has elements >1, and LHTSS corrects both of these deficiencies. It is quite striking how well the LHTSS reconstruction matches the measured reflectance curve of the Munsell chip.

To check just how well the curves produced by LHTSS compare to the reflectance curves measured from 1296 Munsell chip colors, the "reflectance match measure, RMM" was computed and added to the table presented earlier:

| Name | Max $RMM$ | Mean $RMM$ |
|---|---|---|
| **LLS, Linear Least Squares** | 2.46 | 0.88 |
| **LSS, Least Slope Squared** | 1.11 | 0.17 |
| **ILSS, Iterative Least Slope Squared** | 1.04 | 0.16 |
| **LLSS, Least Log Slope Squared** | 0.92 | 0.15 |
| **ILLSS, Iterative Least Log Slope Squared** | 0.86 | 0.15 |
| **LHTSS, Least Hyperbolic Tangent Slope Squared** | 0.84 | 0.14 |

It is evident that LHTSS performs the best of all methods with regard to the comparison to Munsell reflectance curves.

The computational effort required for the LHTSS method is comparable to that of LLSS. The computational effort study performed above, using 140,608 sRGB triplets (every value possible in intervals of five), showed that the LLSS method required 6.77 iterations on average. The LHTSS method requires an average of 5.66 iterations, so its average run times are somewhat smaller (approx 5%).

If it is desired to produce reflectance curves strictly bounded above and below by 1 and 0, the LHTSS method is the best of the group.

---

**Acknowledgments**

This work has been a tremendous learning experience for me, and I want to thank several people who graciously posted high quality material, upon which most of my work has been based: Bruce Lindbloom[11], Bruce MacEvoy[12], Zsolt Kovacs-Vajna[13], David Briggs[14], and Paul Centore[15]. If you are interested in learning more about color theory, these five links are excellent places to start!

**References**

1. Burns, S. A., Subtractive Color Mixture Computation, arXiv:1710.06364 [cs.GR], and http://scottburns.us/subtractive-color-mixture/
2. http://www.cie.co.at/index.php/LEFTMENUE/index.php?i_ca_id=298
3. http://www.w3.org/Graphics/Color/srgb
4. http://www.brucelindbloom.com/index.html?Math.html
5 Cohen J.B., Kappauf W.E. Metameric color stimuli, fundamental metamers, and Wyszecki's metameric blacks. *Am J Psychol* 1982; 95:537–564.
6. Trigt CV. Smoothest reflectance functions. I. Definition and main results. J Opt Soc Am A. 1990;7(10):1891-1904. Trigt CV. Smoothest reflectance functions. II. Complete results. J Opt Soc Am A. 1990;7(12):2208-2222.
7. http://www.munsellcolourscienceforpainters.com/MunsellResources/MunsellResources.html
8. http://scottburns.us/wp-content/uploads/2015/03/X-Rite-2007-Glossy-reflectance-with-D65-sRGB-1485-chips.xlsx
9. https://en.wikipedia.org/wiki/Luminosity_function
10. http://zsolt-kovacs.unibs.it/colormixingtools/cmt-rs2color
11. http://www.brucelindbloom.com/

12. http://www.handprint.com/LS/CVS/color.html
13. http://zsolt-kovacs.unibs.it/colormixingtools
14. http://www.huevaluechroma.com/index.php
15. http://www.munsellcolourscienceforpainters.com/

---

## Appendix: Linked Textual Data

Several data tables and source codes are supplied in the text above via internet links. For archival purposes, these tables and codes are supplied below.

M matrix (3x3)

```
Conversion between tristimulus values, XYZ, and linear rgb, referenced to D65
illuminant

3.243063328, -1.538376194, -0.49893282
-0.968963091, 1.875424508, 0.041543029
0.055683923, -0.204174384, 1.057994536
```

A' matrix (3x36)

```
CIE 1931 color matching functions for 380 to 730 nm by 10 nm intervals

0.001368, 0.004243, 0.01431, 0.04351, 0.13438, 0.2839, 0.34828, 0.3362,
0.2908, 0.19536, 0.09564, 0.03201, 0.0049, 0.0093, 0.06327, 0.1655, 0.2904,
0.43345, 0.5945, 0.7621, 0.9163, 1.0263, 1.0622, 1.0026, 0.85445, 0.6424,
0.4479, 0.2835, 0.1649, 0.0874, 0.04677, 0.0227, 0.011359, 0.00579, 0.002899,
0.00144

0.000039, 0.00012, 0.000396, 0.00121, 0.004, 0.0116, 0.023, 0.038, 0.06,
0.09098, 0.13902, 0.20802, 0.323, 0.503, 0.71, 0.862, 0.954, 0.99495, 0.995,
0.952, 0.87, 0.757, 0.631, 0.503, 0.381, 0.265, 0.175, 0.107, 0.061, 0.032,
0.017, 0.00821, 0.004102, 0.002091, 0.001047, 0.00052

0.00645, 0.02005, 0.06785, 0.2074, 0.6456, 1.3856, 1.74706, 1.77211, 1.6692,
1.28764, 0.81295, 0.46518, 0.272, 0.1582, 0.07825, 0.04216, 0.0203, 0.00875,
0.0039, 0.0021, 0.00165, 0.0011, 0.0008, 0.00034, 0.00019, 0.00005, 0.00002,
0, 0, 0, 0, 0, 0, 0, 0, 0
```

D65 W vector (36x1)

```
Illuminant D65 over 380 to 730 nm in 10 nm intervals

0.499755
0.546482
0.827549
0.91486
0.934318
0.866823
1.04865
1.17008
1.17812
1.14861
1.15923
1.08811
1.09354
1.07802
1.0479
```

```
1.07689
1.04405
1.04046
1.00000
0.963342
0.95788
0.886856
0.900062
0.895991
0.876987
0.832886
0.836992
0.800268
0.802146
0.822778
0.782842
0.697213
0.716091
0.74349
0.61604
0.698856
```

T matrix (3x36)

```
5.47813E-05, 0.000184722, 0.000935514, 0.003096265, 0.009507714, 0.017351596,
0.022073595, 0.016353161, 0.002002407, -0.016177731, -0.033929391, -
0.046158952, -0.06381706, -0.083911194, -0.091832385, -0.08258148, -
0.052950086, -0.012727224, 0.037413037, 0.091701812, 0.147964686,
0.181542886, 0.210684154, 0.210058081, 0.181312094, 0.132064724, 0.093723787,
0.057159281, 0.033469657, 0.018235464, 0.009298756, 0.004023687, 0.002068643,
0.00109484, 0.000454231, 0.000255925

-4.65552E-05, -0.000157894, -0.000806935, -0.002707449, -0.008477628, -
0.016058258, -0.02200529, -0.020027434, -0.011137726, 0.003784809,
0.022138944, 0.038965605, 0.063361718, 0.095981626, 0.126280277, 0.148575844,
0.149044804, 0.14239936, 0.122084916, 0.09544734, 0.067421931, 0.035691251,
0.01313278, -0.002384996, -0.009409573, -0.009888983, -0.008379513, -
0.005606153, -0.003444663, -0.001921041, -0.000995333, -0.000435322, -
0.000224537, -0.000118838, -4.93038E-05, -2.77789E-05

0.00032594, 0.001107914, 0.005677477, 0.01918448, 0.060978641, 0.121348231,
0.184875618, 0.208804428, 0.197318551, 0.147233899, 0.091819086, 0.046485543,
0.022982618, 0.00665036, -0.005816014, -0.012450334, -0.015524259, -
0.016712927, -0.01570093, -0.013647887, -0.011317812, -0.008077223, -
0.005863171, -0.003943485, -0.002490472, -0.001440876, -0.000852895, -
0.000458929, -0.000248389, -0.000129773, -6.41985E-05, -2.71982E-05, -
1.38913E-05, -7.35203E-06, -3.05024E-06, -1.71858E-06
```

$T^{\mathsf{T}}(T\,T^{\mathsf{T}})^{-1}$ matrix (36x3)

```
0.0002, -0.0001, 0.0019
0.0008, -0.0002, 0.0065
0.0042, -0.0009, 0.0334
0.0140, -0.0029, 0.1130
0.0432, -0.0083, 0.3593
```

```
0.0802, -0.0110, 0.7155
0.1063, -0.0000, 1.0915
0.0888, 0.0329, 1.2355
0.0350, 0.0845, 1.1720
-0.0367, 0.1483, 0.8820
-0.1052, 0.2355, 0.5632
-0.1507, 0.3232, 0.3044
-0.2089, 0.4874, 0.1819
-0.2698, 0.7227, 0.1092
-0.2824, 0.9513, 0.0598
-0.2302, 1.1326, 0.0413
-0.1105, 1.1548, 0.0285
0.0473, 1.1288, 0.0227
0.2359, 1.0011, 0.0198
0.4369, 0.8261, 0.0183
0.6450, 0.6415, 0.0176
0.7587, 0.4116, 0.0151
0.8609, 0.2517, 0.0137
0.8476, 0.1274, 0.0114
0.7265, 0.0513, 0.0089
0.5271, 0.0131, 0.0060
0.3731, -0.0016, 0.0041
0.2272, -0.0049, 0.0024
0.1329, -0.0042, 0.0014
0.0724, -0.0026, 0.0008
0.0369, -0.0015, 0.0004
0.0160, -0.0007, 0.0002
0.0082, -0.0004, 0.0001
0.0043, -0.0002, 0.0000
0.0018, -0.0001, 0.0000
0.0010, -0.0000, 0.0000
```

## $B_{12}$ matrix (36x3)

```
Matrix "B12" which converts linear RGB values (0-1) to a
"representative" reflectance curve (over wavelengths
380 to 730 nm, in 10 nm intervals).

0.0933, -0.1729, 1.0796
0.0933, -0.1728, 1.0796
0.0932, -0.1725, 1.0794
0.0927, -0.1710, 1.0783
0.0910, -0.1654, 1.0744
0.0854, -0.1469, 1.0615
0.0723, -0.1031, 1.0308
0.0487, -0.0223, 0.9736
0.0147, 0.0980, 0.8873
-0.0264, 0.2513, 0.7751
-0.0693, 0.4234, 0.6459
-0.1080, 0.5983, 0.5097
-0.1374, 0.7625, 0.3749
-0.1517, 0.9032, 0.2486
-0.1437, 1.0056, 0.1381
-0.1080, 1.0581, 0.0499
-0.0424, 1.0546, -0.0122
0.0501, 0.9985, -0.0487
```

```
0.1641, 0.8972, -0.0613
0.2912, 0.7635, -0.0547
0.4217, 0.6129, -0.0346
0.5455, 0.4616, -0.0071
0.6545, 0.3238, 0.0217
0.7421, 0.2105, 0.0474
0.8064, 0.1262, 0.0675
0.8494, 0.0692, 0.0814
0.8765, 0.0330, 0.0905
0.8922, 0.0121, 0.0957
0.9007, 0.0006, 0.0987
0.9052, -0.0053, 0.1002
0.9073, -0.0082, 0.1009
0.9083, -0.0096, 0.1012
0.9088, -0.0102, 0.1014
0.9090, -0.0105, 0.1015
0.9091, -0.0106, 0.1015
0.9091, -0.0107, 0.1015
```

## LSS (Least Slope Squared) source code (Matlab and Octave)

```
function rho=LSS(B12,sRGB)
% This is the Least Slope Squared (LSS) algorithm for generating
% a "reasonable" reflectance curve from a given sRGB color triplet.
% The reflectance spans the wavelength range 380-730 nm in 10 nm increments.

% It solves min sum(rho_i+1 - rho_i)^2 s.t. T rho = rgb,
% using Lagrangian approach.

% B12 is upper-right 36x3 part of inv([D,T';T,zeros(3)])
% sRGB is a three-element vector of target D65-referenced sRGB values
%      in 0-255 range,
% rho is a 36x1 vector of reflectance values over wavelengths 380-730 nm,

% Written by Scott Allen Burns, 4/25/15.
% Licensed under a Creative Commons Attribution-ShareAlike 4.0 International
% License (http://creativecommons.org/licenses/by-sa/4.0/).
% For more information, see
% http://scottburns.us/reflectance-curves-from-srgb/

% compute target linear rgb values
sRGB=sRGB(:)/255; % convert to 0-1 column vector
rgb=zeros(3,1);
% remove gamma correction to get linear rgb
for i=1:3
    if sRGB(i)<0.04045
        rgb(i)=sRGB(i)/12.92;
    else
        rgb(i)=((sRGB(i)+0.055)/1.055)^2.4;
    end
end

% matrix multiply
rho=B12*rgb;
```

## LLSS (Least Log Slope Squared) source code (Matlab and Octave)

```
function rho=LLSS(T,sRGB)
% This is the Least Log Slope Squared (LLSS) algorithm for generating
% a "reasonable" reflectance curve from a given sRGB color triplet.
% The reflectance spans the wavelength range 380-730 nm in 10 nm increments.

% Solves min sum(z_i+1 - z_i)^2 s.t. T exp(z) = rgb, where
% z=log(reflectance), using Lagrangian formulation and Newton's method.
% Allows reflectance values >1 to be in solution.

% T is 3x36 matrix converting reflectance to D65-weighted linear rgb,
% sRGB is a 3 element vector of target D65 referenced sRGB values (0-255),
% rho is a 36x1 vector of reconstructed reflectance values, all > 0,

% For more information, see
% http://scottburns.us/reflectance-curves-from-srgb/
% Written by Scott Allen Burns, March 2015.
% Licensed under a Creative Commons Attribution-ShareAlike 4.0 International
% License (http://creativecommons.org/licenses/by-sa/4.0/).

% initialize outputs to zeros
rho=zeros(36,1);

% handle special case of (0,0,0)
if all(sRGB==0)
    rho=0.0001*ones(36,1);
    return
end

% 36x36 difference matrix for Jacobian
% having 4 on main diagonal and -2 on off diagonals,
% except first and last main diagonal are 2.
D=full(gallery('tridiag',36,-2,4,-2));
D(1,1)=2;
D(36,36)=2;

% compute target linear rgb values
sRGB=sRGB(:)/255; % convert to 0-1
rgb=zeros(3,1);
% remove gamma correction to get linear rgb
for i=1:3
    if sRGB(i)<0.04045
        rgb(i)=sRGB(i)/12.92;
    else
        rgb(i)=((sRGB(i)+0.055)/1.055)^2.4;
    end
end

% initialize
z=zeros(36,1); % starting point all zeros
lambda=zeros(3,1); % starting Lagrange mult
maxit=100; % max number of iterations
ftol=1.0e-8; % function solution tolerance
deltatol=1.0e-8; % change in oper pt tolerance
count=0; % iteration counter

% Newton's method iteration
while count <= maxit
```

```
    r=exp(z);
    v=-diag(r)*T'*lambda; % 36x1
    m1=-T*r; % 3x1
    m2=-T*diag(r); % 3x36
    F=[D*z+v;m1+rgb]; % 39x1 function vector
    J=[D+diag(v),m2';m2,zeros(3)]; % 39x39 Jacobian matrix
    delta=J\(-F); % solve Newton system of equations J*delta = -F
    z=z+delta(1:36); % update z
    lambda=lambda+delta(37:39); % update lambda
    if all(abs(F)<ftol) % check if functions satisfied
        if all(abs(delta)<deltatol) % check if variables converged
            % solution found
            disp(['Solution found after ',num2str(count),' iterations'])
            rho=exp(z);
            return
        end
    end
    count=count+1;
end
disp(['No solution found in ',num2str(maxit),' iterations.'])
```

## ILLSS (Iterative Least Log Slope Squared) source code (Matlab and Octave)

```
function rho=ILLSS(T,sRGB)
% This is the Iterative Least Log Slope Squared (ILLSS) algorithm for
% generating a "reasonable" reflectance curve from a given sRGB color
% triplet. The reflectance spans the wavelength range 380-730 nm in 10 nm
% increments.

% It solves min sum(z_i+1 - z_i)^2 s.t. T exp(z) = rgb, K z = 0, where
% z=log(reflectance), using Lagrangian approach and Newton's method.
% Clips values >1 and repeats optimization until all reflectance <=1.

% T    is 3x36 matrix converting reflectance to linear rgb over the
%      range 380-730 nm,
% sRGB is a 3 element vector of target D65 referenced sRGB values
%      in 0-255 range,
% rho  is a 36x1 vector of reflectance values (0->1] over
%      wavelengths 380-730 nm,

% Written by Scott Allen Burns, 4/11/15.
% Licensed under a Creative Commons Attribution-ShareAlike 4.0 International
% License (http://creativecommons.org/licenses/by-sa/4.0/).
% For more information, see
% http://scottburns.us/reflectance-curves-from-srgb/

% initialize output to zeros
rho=zeros(36,1);

% handle special case of (0,0,0)
if all(sRGB==0)
    rho=0.0001*ones(36,1);
    return
end

% handle special case of (255,255,255)
```

```
if all(sRGB==255)
    rho=ones(36,1);
    return
end

% 36x36 difference matrix having 4 on main diagonal and -2 on off diagonals,
% except first and last main diagonal are 2.
D=full(gallery('tridiag',36,-2,4,-2));
D(1,1)=2;
D(36,36)=2;

% compute target linear rgb values
sRGB=sRGB(:)/255; % convert to 0-1 column vector
rgb=zeros(3,1);
% remove gamma correction to get linear rgb
for i=1:3
    if sRGB(i)<0.04045
        rgb(i)=sRGB(i)/12.92;
    else
        rgb(i)=((sRGB(i)+0.055)/1.055)^2.4;
    end
end

% outer iteration to get all refl <=1
maxouter=10;
outer_count=0; % counter for outer iteration
while (any(rho>1) && outer_count<=maxouter) || all(rho==0)
    % create K matrix for fixed refl constraints
    fixed_refl=find(rho>=1)';
    numfixed=length(fixed_refl);
    K=zeros(numfixed,36);
    for i=1:numfixed
        K(i,fixed_refl(i))=1;
    end

    % initialize
    z=zeros(36,1); % starting point all zeros
    lambda=zeros(3,1); % starting point for lambda
    mu=zeros(numfixed,1); % starting point for mu
    maxit=50; % max number of iterations
    ftol=1.0e-8; % function solution tolerance
    deltatol=1.0e-8; % change in oper pt tolerance
    count=0; % iteration counter

    % Newton's method iteration
    while count <= maxit
        r=exp(z);
        v=-diag(r)*T'*lambda; % 36x1
        m1=-T*r; % 3x1
        m2=-T*diag(r); % 3x36
        F=[D*z+v+K'*mu;m1+rgb;K*z]; % function vector
        J=[D+diag(v),[m2',K'];[m2;K],zeros(numfixed+3)]; % Jacobian matrix
        delta=J\(-F); % solve Newton system of equations J*delta = -F
        z=z+delta(1:36); % update z
        lambda=lambda+delta(37:39); % update lambda
        mu=mu+delta(40:end);
        if all(abs(F)<ftol) % check if functions satisfied
```

```
            if all(abs(delta)<deltatol) % check if variables converged
                % solution found
                disp(['Inner loop solution found after ',num2str(count),...
                      ' iterations'])
                rho=exp(z);
                break
            end
        end
        count=count+1;
    end
    if count>=maxit
        disp(['No inner loop solution found after ',num2str(maxit),...
             ' iterations.'])
    end
    outer_count=outer_count+1;
end
if outer_count<maxouter
    disp(['Outer loop solution found after ',num2str(outer_count),...
          ' iterations'])
else
    disp(['No outer loop solution found after ',num2str(maxouter),...
          ' iterations.'])
end
```

## ILSS (Iterative Least Slope Squared) source code (Matlab and Octave)

```
function rho=ILSS(B11,B12,sRGB)
% This is the Iterative Least Slope Squared (ILSS) algorithm for generating
% a "reasonable" reflectance curve from a given sRGB color triplet.
% The reflectance spans the wavelength range 380-730 nm in 10 nm increments.

% It solves
% min  sum(rho_i+1 - rho_i)^2
% s.t. T rho = rgb,
%      K1 rho = 1,
%      K0 rho = 0,
% using Lagrangian formulation and iteration to keep all rho (0-1].

% B11  is upper-left 36x36 part of inv([D,T';T,zeros(3)])
% B12  is upper-right 36x3 part of inv([D,T';T,zeros(3)])
% sRGB is a 3-element vector of target D65-referenced sRGB values (0-255),
% rho  is a 36x1 vector of reflectance values (0->1] over
%      wavelengths 380-730 nm,

% Written by Scott Allen Burns, 4/26/15.
% Licensed under a Creative Commons Attribution-ShareAlike 4.0 International
% License (http://creativecommons.org/licenses/by-sa/4.0/).
% For more information, see
% http://scottburns.us/reflectance-curves-from-srgb/

rho=ones(36,1)/2; % initialize output to 0.5
rhomin=0.00001; % smallest refl value

% handle special case of (255,255,255)
if all(sRGB==255)
    rho=ones(36,1);
```

```
    return
end

% handle special case of (0,0,0)
if all(sRGB==0)
    rho=rhomin*ones(36,1);
    return
end

% compute target linear rgb values
sRGB=sRGB(:)/255; % convert to 0-1 column vector
rgb=zeros(3,1);
% remove gamma correction to get linear rgb
for i=1:3
    if sRGB(i)<0.04045
        rgb(i)=sRGB(i)/12.92;
    else
        rgb(i)=((sRGB(i)+0.055)/1.055)^2.4;
    end
end

R=B12*rgb;

% iteration to get all refl 0-1
maxit=10; % max iterations
count=0; % counter for iteration
while ( (any(rho>1) || any(rho<rhomin)) && count<=maxit ) || count==0
    % create K1 matrix for fixed refl at 1
    fixed_upper_logical = rho>=1;
    fixed_upper=find(fixed_upper_logical);
    num_upper=length(fixed_upper);
    K1=zeros(num_upper,36);
    for i=1:num_upper
        K1(i,fixed_upper(i))=1;
    end

    % create K0 matrix for fixed refl at rhomin
    fixed_lower_logical = rho<=rhomin;
    fixed_lower=find(fixed_lower_logical);
    num_lower=length(fixed_lower);
    K0=zeros(num_lower,36);
    for i=1:num_lower
        K0(i,fixed_lower(i))=1;
    end

    % set up linear system
    K=[K1;K0];
    C=B11*K'/(K*B11*K'); % M*K'*inv(K*M*K')
    rho=R-C*(K*R-[ones(num_upper,1);rhomin*ones(num_lower,1)]);
    rho(fixed_upper_logical)=1; % eliminate FP noise
    rho(fixed_lower_logical)=rhomin; % eliminate FP noise

    count=count+1;
end
if count>=maxit
    disp(['No solution found after ',num2str(maxit),' iterations.'])
end
```

LHTSS (Least Hyperbolic Tangent Slope Squared) source code (Matlab and Octave)

```
function rho=LHTSS(T,sRGB)
% This is the Least Hyperbolic Tangent Slope Squared (LHTSS) algorithm for
% generating a "reasonable" reflectance curve from a given sRGB color tri-
plet.
% The reflectance spans the wavelength range 380-730 nm in 10 nm increments.

% Solves min sum(z_i+1 - z_i)^2 s.t. T ((tanh(z)+1)/2) = rgb,
% using Lagrangian formulation and Newton's method.
% Reflectance will always be in the open interval (0->1).

% T is 3x36 matrix converting reflectance to D65-weighted linear rgb,
% sRGB is a 3 element vector of target D65 referenced sRGB values in 0-255
range,
% rho is a 36x1 vector of reconstructed reflectance values, all (0->1),

% Written by Scott Allen Burns, May 2019.
% Licensed under a Creative Commons Attribution-ShareAlike 4.0 International
% License (http://creativecommons.org/licenses/by-sa/4.0/).
% For more information, see http://scottburns.us/reflectance-curves-from-
srgb/

% initialize outputs to zeros
rho=zeros(36,1);

% handle special case of (0,0,0)
if all(sRGB==0)
    rho=0.0001*ones(36,1);
    return
end

% handle special case of (255,255,255)
if all(sRGB==255)
    rho=ones(36,1);
    return
end

% 36x36 difference matrix for Jacobian
% having 4 on main diagonal and -2 on off diagonals,
% except first and last main diagonal are 2.
D=full(gallery('tridiag',36,-2,4,-2));
D(1,1)=2;
D(36,36)=2;

% compute target linear rgb values
sRGB=sRGB(:)/255; % convert to 0-1
rgb=zeros(3,1);
% remove gamma correction to get linear rgb
for i=1:3
    if sRGB(i)<0.04045
        rgb(i)=sRGB(i)/12.92;
    else
        rgb(i)=((sRGB(i)+0.055)/1.055)^2.4;
    end
end
```

```
% initialize
z=zeros(36,1); % starting point all rho=1/2
lambda=zeros(3,1); % starting Lagrange mult
maxit=100; % max number of iterations
ftol=1.0e-8; % function solution tolerance
count=0; % iteration counter

% Newton's method iteration
while count <= maxit
    d0 = (tanh(z) + 1)/2;
    d1 = diag((sech(z).^2)/2);
    d2 = diag(-sech(z).^2.*tanh(z));
    F = [D*z + d1*T'*lambda; T*d0 - rgb]; % 39x1 F vector
    J = [D + diag(d2*T'*lambda), d1*T'; T*d1, zeros(3)]; % 39x39 J matrix
    delta=J\(-F); % solve Newton system of equations J*delta = -F
    z=z+delta(1:36); % update z
    lambda=lambda+delta(37:39); % update lambda
    if all(abs(F)<ftol)
        % solution found
        rho=(tanh(z)+1)/2;
        return
    end
    count=count+1;
end
disp(['No solution found in ',num2str(maxit),' iterations.'])
```